\documentclass{article} 
\usepackage{colm2024_conference}

\usepackage{graphicx}
\usepackage{subfigure}
\usepackage{wrapfig}
\usepackage{multirow}
\usepackage{tikz}
\usetikzlibrary{
    arrows,
    arrows.meta,
    shapes, shapes.multipart,
    decorations.pathmorphing,
    decorations.markings,
    backgrounds,
    calc,
    positioning,
    fit,
}
\usepackage{amsmath} 

\usepackage{microtype}
\usepackage{hyperref}
\usepackage{makecell}
\usepackage{xcolor,soul}
\usepackage{adjustbox}
\usepackage{array}
\usepackage{url}
\usepackage{booktabs}
\usepackage{authblk}
\definecolor{darkblue}{rgb}{0.4, 0.7, 0.9}
\definecolor{darkgreen}{rgb}{0.3,0.9, 0.7}
\definecolor{darkred}{rgb}{0.9, 0.8, 0.5}
\definecolor{lightblue}{rgb}{.70,.75,1}
\definecolor{lightgreen}{rgb}{.70,1,0.75}
\definecolor{lightpink}{rgb}{.97,0.55,0.71}
\definecolor{lightyellow}{rgb}{1,0.93,0.32}
\hypersetup{colorlinks=true, citecolor=darkblue, linkcolor=darkblue, urlcolor=darkblue}

\usepackage{svg}

\title{BindGPT: A Scalable Framework for 3D Molecular Design via Language Modeling and Reinforcement Learning}

\author[1,3,4]{Artem Zholus}
\author[1]{Maksim Kuznetsov}
\author[2]{Roman Schutski}
\author[1]{Rim Shayakhmetov}
\author[1]{Daniil Polykovskiy}
\author[3,4,5]{Sarath Chandar}
\author[2]{Alex Zhavoronkov}
\affil[1]{Insilico Medicine Canada Inc.}
\affil[2]{Insilico Medicine AI Limited}
\affil[3]{Mila – Quebec AI Institute}
\affil[4]{Polytechnique Montréal}
\affil[5]{CIFAR AI Chair}

\newcommand\blfootnote[1]{%
  \begingroup
  \renewcommand\thefootnote{}\footnote{#1}%
  \addtocounter{footnote}{-1}%
  \endgroup
}
\colmfinalcopy 
\begin{document}

\maketitle

\begin{abstract}
Generating novel active molecules for a given protein is an extremely challenging task for generative models that requires an understanding of the complex physical interactions between the molecule and its environment. In this paper, we present a novel generative model, BindGPT which uses a conceptually simple but powerful approach to create 3D molecules within the protein's binding site. Our model produces molecular graphs and conformations jointly, eliminating the need for an extra graph reconstruction step. We pretrain BindGPT on a large-scale dataset and fine-tune it with reinforcement learning using scores from external simulation software. We demonstrate how a single pretrained language model can serve at the same time as a 3D molecular generative model, conformer generator conditioned on the molecular graph, and a pocket-conditioned 3D molecule generator. Notably, the model does not make any representational equivariance assumptions about the domain of generation. We show how such simple conceptual approach combined with pretraining and scaling can perform on par or better than the current best specialized diffusion models, language models, and graph neural networks while being two orders of magnitude cheaper to sample.

\end{abstract}

\section{Introduction}
\label{submission}

The landscape of drug discovery presents immense challenges and risks, demanding substantial investments of time and resources to design, test, and deliver new medicines to the market. Within this context, Computer-Aided Drug Design (CADD) \citep{yu2017cadd} stands as a pivotal methodology, harnessing software screenings and physical simulations to facilitate a more efficient exploration of the vast space of drug-like molecules, estimated to be around $10^{60}$ in size \citep{Polishchuk2013-fp,GomezBombarelli2018}. Deep learning advancements have revolutionized this exploration by leveraging neural generative models trained on extensive compound datasets. Notably, the textual representation of molecular structures using SMILES \citep{Weininger1988} and SELFIES \cite{krenn2020selfies} has enabled the utilization of Language Models for the generation of novel, drug-like molecular compounds \citep{Segler2018,Bagal2022}.
\blfootnote{Correspondence to: \href{mailto:artem.zholus@mila.quebec}{\texttt{artem.zholus@mila.quebec}}, \href{mailto:r.schutski@insilicomedicine.com}{\texttt{r.schutski@insilicomedicine.com}}}

Recent research has demonstrated the capability of deep generative models to generate novel molecular compounds directly in 3D, with the flexibility to incorporate protein pocket and ligand subfragment conditions. Among these, diffusion models such as EDM \citep{hoogeboom2022edm} and DiffDock \citep{corsoDiffDockDiffusionSteps2023} initiate the generation process with an arbitrary spatial distribution of atoms and progressively refine their positions to yield physically viable molecular structures. Meanwhile, autoregressive models like Pocket2Mol \citep{peng2022pocket2mol} sequentially predict the type and location of each successive atom, building upon the existing molecular framework. Additionally, work by \citet{flamshepherd2023language} has highlighted the proficiency of language models in handling spatial representations of molecular and protein structures through formats like XYZ, CIF, and PDB. However, it's noteworthy that most spatial molecular generators focus exclusively on atom types and locations. They depend on supplementary tools, such as OpenBabel \citep{oboyle2011openbabel}, for the critical task of bond reconstruction. This reliance can introduce vulnerabilities, as the precision required for atom placement means that minor positional adjustments can significantly alter reconstructed molecular bonds or even make the molecular graph disconnected.

\begin{figure*}[t!]
    \centering
    \resizebox{\textwidth}{!}{
        \input{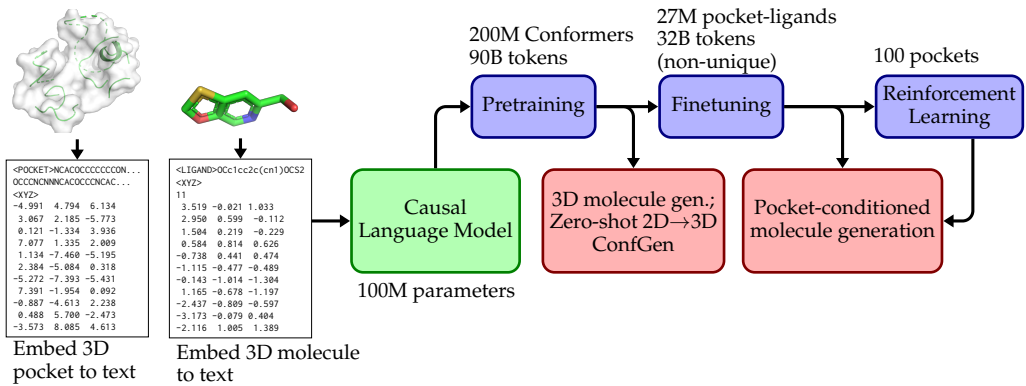}
    }
    \vspace{-10pt}
    \caption{The pipeline of our 3D molecule pretraining-finetuning paradigm. During pretraining the model is trained on a mix of molecules and pockets in isolation. But at finetuning, we simply concatenate the pocket text representation and the molecule text representation for each pocket-ligand pair.}
    \label{fig:method_scheme}
    \vspace{-15pt}
\end{figure*}

In this work, we introduce a novel framework that applies language modeling to the domain of 3D molecular data represented by textual tokens. This entirely data-driven approach, devoid of any inductive biases at both the model and representation levels, capitalizes on the established GPT paradigm, integrating cutting-edge techniques to enhance the scalability of model training and inference. By adopting the language model pretraining paradigm, our framework showcases the ability to foster a powerful causal language model adept at navigating the complex space of 3D molecules. This proficiency is demonstrated through successful applications in downstream tasks, including learning the distribution of 3D molecules, generating 3D conformations and the generation of molecules with targeted binding affinity to specific proteins.

Our main contributions are the following:

\begin{itemize}
    \item We introduce BindGPT, a Language Model for handling spatial molecular structures in text format. It uses structural SMILES and spatial XYZ formats to describe molecular graphs and atom locations, eliminating the dependency on external software for graph reconstruction.
    \item We propose scalable pretraining-finetuning method for drug discovery in 3D that covers several 3D molecular generation tasks in a single paradigm. 
    \item We show how BindGPT can create accurate and realistic 3D molecular structures both zero-shot and after finetuning, with the option to include molecular graphs or protein pocket descriptions as prompts. The method offers comparable generation quality to leading approaches with the speedup of up to $100$x.
    \item Finally, we demonstrate the effectiveness of the Reinforcement Learning framework to finetune BindGPT with an external feedback from docking software. We show that the resulting model can find structures with high binding scores for any given protein as a result for the RL finetuning.

\end{itemize}
\section{Background}
\label{background}

\vspace{-15pt}
\textbf{Molecule Generation} Small drug-like molecules can be represented as 2D or 3D graphs with node and edge attributes. However, one of the most popular molecular representation in the machine learning community is SMILES \citep{Weininger1988}, which can be seen as a compressed textual encoding of the Depth-First-Search applied to the molecular graph. It's simplicity and expressivity made it work very well with language models - even a simple LSTM \citep{hochreiter1997long} model can outperform graph-neural networks for the molecule generation task \citep{FlamShepherd2022}. In addition, SELFIES \citep{Krenn_2022}, is a modification of SMILES which is a robust string representation such that every SELFIES token sequence is a valid molecule and vice versa.

The biological function of small molecules arises through their binding to specific protein pockets. The spatial structure of the protein pocket is an essential domain knowledge to increase the efficiency of molecular generation in drug design tasks. With the increase of molecular structure datasets sizes \citep{Francoeur2020,hu2005binding} a plethora of pocket-conditioned generators emerged \citep{peng2022pocket2mol,luo20223d,linDiffBPGenerativeDiffusion2022, corsoDiffDockDiffusionSteps2023}. The challenge with pocket-conditioned molecular generation arises from a relatively small size of existing 3D binding poses datasets, which motivated a heavy use of specialized architectures, like SE(3) equivariant neural networks \citep{hoogeboom2022edm}. 

\textbf{Molecular Generative Models in 3D.}  Apart from Language Models, there exist other types of generative models that approach molecule generation, including 3D-aware ones. The first group of works uses Diffusion Models, which employ the denoising diffusion process \citep{ho2020denoising,song2021scorebased} to learn to recover the data from noise. The second group of works relies on Graph Neural Networks (GNNs) to autoregressively build 2D or 3D molecular graphs.  These approaches can be combined as GNNs can serve as efficient backbones for diffusion process once they are node-equivariant \citep{niu2020permutation} (to generate 2D graphs) or SE(3)-equivariant \citep{pmlr-v202-peng23b} (to generate 3D graphs). SBDD \citep{luo20223d} model uses autoregressive graph generation for pocket-conditioned molecule generation. TargetDiff \citep{schneuing2023structurebased} generalizes this model to use a diffusion-GNN for the same task. Pocket2Mol \citep{peng2022pocket2mol} uses an informed autoregressive sampling mechanism for efficient pocket-conditioned molecule generation. 
Another batch of works use the aforementioned methods for the unconditional molecule generation. EDM \citep{hoogeboom2022edm} proposes an E(3) equivariant diffusion model for molecule generation. MolDiff \citep{pmlr-v202-peng23b} is a diffusion model that addresses the inconsistency problem between generated atoms and bonds connecting them.

\textbf{Language Models for Drug Discovery.} The language models show outstanding results in drug discovery domain. The molecular structures can be easily represented in a textual formats like SMILES \citep{Segler2018} or SELFIES \citep{FlamShepherd2022}, enabling the effective training of well-known language model architectures on large datasets of chemical entities. Recent studies reveals the potential of applying language models to address various challenges in drug discovery.
For instance, LigGPT \citep{Bagal2022} leverages the GPT \citep{radford2018improving} architecture to generate molecular structures given the conditions of molecular descriptors.
MoLFormer \citep{katharopoulos_et_al_2020} incorporates billion-size chemical entities database to perform large-scale pretrain and further finetune to predict molecular properties. BARTSmiles \citep{chilingaryan2022bartsmiles} is developed atop of BART \citep{lewis2019bart} architecture, training a meaningful chemical representation and refining it for tasks such as chemical property prediction, chemical reaction prediction, and retrosynthesis. 

However, the challenge of 3D molecule generation has received limited attention in language model approach. Three notable studies in this area include the XYZ-transformer \citep{flamshepherd2023language}, Uni-Mol \citep{zhou2023unimol}, and Lingo3DMol \citep{Feng2024}. The XYZ-transformer leverages GPT for generating atom-wise description of molecular and protein structures. 
Uni-Mol modifies BERT for large-scale pretraining on a large 3D structures dataset. In particular, Uni-Mol formulates molecular tasks as coordinates-to-coordinates mapping given molecular graph. 
\textbf{To the best of our knowledge, we propose the first approach that applies the modern decoder-only language modeling paradigm to the 3D drug discovery problem with several downstream applications.} 
\section{Method}

The key idea of our method is utilizing an autoregressive token generation model, influenced by GPT-based models, to solve several 3D small molecule generation tasks in one simple yet flexible paradigm. 
The main principle in our approach is to formulate several 3D molecular design task as prompted generation of text. To achieve that, we layout the tokens of a condition before the tokens of the object to generate. For instance, a prompt can be the protein pocket for the pocket-conditioned generation task or the 2D molecular structure for the conformation generation task.

\subsection{Architecture and text format}
\label{s:arm}

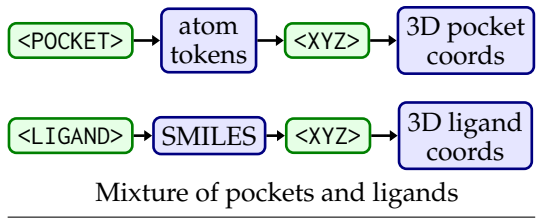
\begin{wrapfigure}{r}{.55\textwidth}
\vspace{-0.6cm}
\begin{center}
\begin{tikzpicture}[
    scale=1,
    block/.style={draw, rectangle,anchor=center,align=center, rounded corners=1mm,line width=1pt},
    arr/.style={-{Triangle[width=4pt,length=3pt]}, line width=1pt}
]
    \node[block,
          draw=green!50!black,
          fill=green!10!white
    ] (pock) {\texttt{<POCKET>}};
    \node[block,
          right=1em of pock,
          draw=blue!50!black,
          fill=blue!10!white
    ] (pockd) {atom \\ tokens};
    \draw[arr] (pock) -- (pockd);
    \node[block,
          right=1em of pockd,
          draw=green!50!black,
          fill=green!10!white
    ] (pxyz) {\texttt{<XYZ>}};
    \draw[arr] (pockd) -- (pxyz);
    \node[block,
          right=1em of pxyz,
          draw=blue!50!black,
          fill=blue!10!white
    ] (pxyzd) {3D pocket \\ coords};

    \node[block,
          below=2.2em of pock,
          draw=green!50!black,
          fill=green!10!white
    ] (lig) {\texttt{<LIGAND>}};    
    \draw[arr] (pxyz) -- (pxyzd);
    
    \node[block,
          right=0.7em of lig,
          draw=blue!50!black,
          fill=blue!10!white
    ] (smi) {SMILES};
    \draw[arr] (lig) -- (smi);
    \node[block,
          right=0.7em of smi,
          draw=green!50!black,
          fill=green!10!white
    ] (lxyz) {\texttt{<XYZ>}};
    \draw[arr] (smi) -- (lxyz);
    \node[block,
          right=1em of lxyz,
          draw=blue!50!black,
          fill=blue!10!white
    ] (lxyzd) {3D ligand \\ coords};
    \draw[arr] (lxyz) -- (lxyzd);

    \node[below left=1.5em and 2.5em of smi,anchor=west] (task1) {Mixture of pockets and ligands};
    \draw[-] (current bounding box.south east) -- (current bounding box.south west);

\end{tikzpicture}
\end{center}
\vspace{-0.5cm}
\caption{Data layout during the pretraining. Arrows show the tokens sequence order. Nodes such as \texttt{<POCKET>}
 show special tokens. Training is done on a mixture of pocket and ligand datasets. Concrete examples are shown in Appendix \ref{appe:tokenization}.}
\vspace{-0.4cm}
\label{fig:layout_pretrain}
\end{wrapfigure}

In our work we follow the decoder-only paradigm in Language Models and use the GPT-NeoX\footnote{Despite other architectures existing today (such as LLaMa2 \citep{touvron2023llama}), our main focus was building a general framework of working with 3D molecules that can work with any Language Model architecture.} architecture \citep{gpt-neox-20b} that utilizes rotary position embeddings \citep{rope}. This technique allows for the length generalization, which is required since the sequence lengths may vary significantly between the pretraining, and fine-tuning stages. 

Similarly to \cite{flamshepherd2023language}, we use the XYZ representation as a base format to describe the spatial atom allocation. The idea of XYZ format is to represent the atom type and its 3D coordinates within every line on text. The main drawback of this format is the lack of charge and connectivity information. One should use external software like RDKit \citep{rdkit} or OpenBabel \citep{oboyle2011openbabel} to reconstruct the molecular graph. It introduces an instability since even small noise in atom positions can drastically change the reconstructed graph or even break it down \citep{moldiff}. To alleviate that, we propose to couple the XYZ format with the SMILES format. The latter can efficiently represent the molecular structure, while the former allows describing atom positions. To align these two formats we enforce to have the same atom ordering in both. We also remove the atom symbol from the XYZ representation as it already was shown in SMILES. For proteins, there is no need to describe their connectivity, therefore we simply write atom names grouped by aminoacids.

A schematic example of the two kind of the model input is shown in Figure \ref{fig:layout_pretrain} and the very detailed example of concrete input sequences (including its tokenization) is shown in Figure \ref{fig:example_tokens} in Appendix \ref{appe:tokenization}. In particular, the sequence starts with the \texttt{<LIGAND>} token followed by a SMILES string, tokenized at the character level. Next, there goes the \texttt{<XYZ>} special token marking the end of SMILES and the beginning of the coordinate part of the string. The tokenization strategy uses 6 tokens per 3D position: we use one token for the integer part and one token for the fractional part of the number. When working with protein pockets, we use a similar strategy. Specifically, the sequence begins with the \texttt{<POCKET>} token followed by the sequence of atoms where each atom is a separate token. Since pockets can be hundreds of atoms large, we follow the AlphaFold's \citep{jumper2021highly} approach and retain only the 3D coordinates of the Alpha-carbon atoms in the corresponding aminoacids. An example of the final representation of pockets is shown in Figure \ref{fig:example_pockets}.

\subsection{Pretraining}
\label{sec:pretrain}

In this work, we aim to leverage insights accumulated by the NLP community in the paradigm of Large Language Models: pretraining-finetuning, prompting, scaling, finetuning with Reinforcement Learning, tool use, etc.\citep{kaplan2020scaling,hoffmann2022training,radford2019language}. Since our model covers only a specialized domain of molecular tasks, it does not require trillion-scale diverse datasets for good performance as NLP tasks do. Thus, we use a large-scale but specialized dataset of 3D molecules and protein pockets. During pretraining, we use the model with 108M parameters consisting of 15 layers, 12 heads, and a hidden size of 768. We found this size of the model to be enough for the tasks we care about - generating molecules in 3D (See Appendix \ref{app:scaling} for a justification of the size).  Every sequence in the training batch is either a ligand sequence of tokens or a pocket sequence of tokens following the scheme described earlier. Since the dataset has much fewer pockets than ligands, for one epoch of training on ligands, we do 5 epochs of training on proteins, that is, around 8\% of all tokens seen by the model are pocket tokens. To speedup and stabilize pretraining, we use large batch training \citep{keskar2017on} with 1.6M tokens per one training step. We found this many tokens per batch to be important for stable training in this task even with smaller learning rates. The detailed description of the training implementation is provided in Appendix \ref{app:efficient_training}. Despite the wide use of transformers in drug discovery, the majority of current works in this space do not use recent advancements of efficient Language Models pretraining: neither technical ones, such as Flash-attention \citep{dao2023flashattention2} or DeepSpeed \citep{deepspeed}, nor the algorithimic ones, such as learning rate scaling. Our work aims to fill this gap by demonstrating the effectiveness of the pretraining for 3D drug discovery.

\subsection{Finetuning}
\label{sec:ft}

\subsubsection{Supervised finetuning} 
\label{sec:sft}

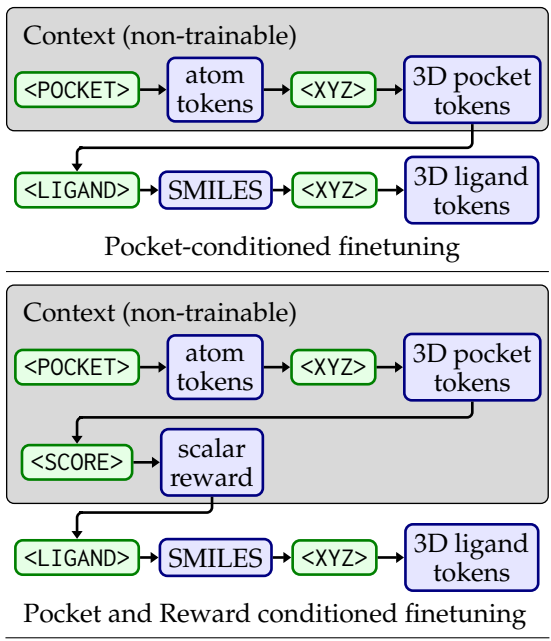
\begin{wrapfigure}{r}{.55\textwidth}
\vspace{-1.6cm}
\begin{center}
\begin{tikzpicture}[
    scale=1,
    wrapper/.style 2 args={%
      local bounding box=localbb,
      execute at end scope={ 
      \begin{scope}[on background layer] 
      \node[inner sep = #1,
            draw,
            fit=(localbb),
            fill = gray!30,
            rounded corners=1mm,line width=0.5pt,
            label = {above:#2}
            ]{};  
      \end{scope}}},
    block/.style={draw, rectangle,anchor=center,align=center, rounded corners=1mm,line width=1pt},
    arr/.style={-{Triangle[width=4pt,length=3pt]}, line width=1pt}
]   
    \begin{scope}[wrapper={0.3em}{}]
    \node[block,
          draw=green!50!black,
          fill=green!10!white
    ] (pock) {\texttt{<POCKET>}};
    \node[above left=1.3em and 0em of pock, anchor=west] (ctx_lab) {Context (non-trainable)};
    \node[block,
          right=1em of pock,
          draw=blue!50!black,
          fill=blue!10!white
    ] (pockd) {atom \\ tokens};
    \draw[arr] (pock) -- (pockd);
    \node[block,
          right=1em of pockd,
          draw=green!50!black,
          fill=green!10!white
    ] (pxyz) {\texttt{<XYZ>}};
    \draw[arr] (pockd) -- (pxyz);
    \node[block,
          right=1em of pxyz,
          draw=blue!50!black,
          fill=blue!10!white
    ] (pxyzd) {3D pocket \\ tokens};
    \end{scope}

    \node[block,
          below=2.4em of pock,
          draw=green!50!black,
          fill=green!10!white
    ] (lig) {\texttt{<LIGAND>}};    
    \draw[arr] (pxyz) -- (pxyzd);
    \draw[arr, rounded corners=2pt] (pxyzd.south) |- ($(pock.south) - (-0.1,1.5em)$) -|  (lig.north);
    
    \node[block,
          right=0.7em of lig,
          draw=blue!50!black,
          fill=blue!10!white
    ] (smi) {SMILES};
    \draw[arr] (lig) -- (smi);
    \node[block,
          right=0.7em of smi,
          draw=green!50!black,
          fill=green!10!white
    ] (lxyz) {\texttt{<XYZ>}};
    \draw[arr] (smi) -- (lxyz);
    \node[block,
          right=1em of lxyz,
          draw=blue!50!black,
          fill=blue!10!white
    ] (lxyzd) {3D ligand \\ tokens};
    \draw[arr] (lxyz) -- (lxyzd);

    \node[below left=1.5em and 2.4em of smi,anchor=west] (task1) {Pocket-conditioned finetuning};
    \draw[-] (current bounding box.south east) -- (current bounding box.south west);

    \begin{scope}[wrapper={0.3em}{}]
    \node[block,
          below=5.3em of lig,
          draw=green!50!black,
          fill=green!10!white
    ] (pock2) {\texttt{<POCKET>}};
    \node[above left=1.3em and 0em of pock2, anchor=west] (ctx_lab2) {Context (non-trainable)};
    \node[block,
          right=1em of pock2,
          draw=blue!50!black,
          fill=blue!10!white
    ] (pockd2) {atom \\ tokens};
    \draw[arr] (pock2) -- (pockd2);
    \node[block,
          right=1em of pockd2,
          draw=green!50!black,
          fill=green!10!white
    ] (pxyz2) {\texttt{<XYZ>}};
    \draw[arr] (pockd2) -- (pxyz2);
    \node[block,
          right=1em of pxyz2,
          draw=blue!50!black,
          fill=blue!10!white
    ] (pxyzd2) {3D pocket \\ tokens};

    \node[block,
          below=2.2em of pock2,
          draw=green!50!black,
          fill=green!10!white
    ] (scr2) {\texttt{<SCORE>}};  

    \draw[arr, rounded corners=2pt] (pxyzd2.south) |- ($(pock2.south) - (-0.1,1.2em)$) -|  (scr2.north);

    \node[block,
          right=1em of scr2,
          draw=blue!50!black,
          fill=blue!10!white
    ] (scrd2) {scalar \\ reward};
    \draw[arr] (scr2) -- (scrd2);
    \end{scope}

    \node[block,
          below=2.2em of scr2,
          draw=green!50!black,
          fill=green!10!white
    ] (lig2) {\texttt{<LIGAND>}};    
    \draw[arr] (pxyz2) -- (pxyzd2);
    \draw[arr, rounded corners=2pt] (scrd2.south) |- ($(scr2.south) - (-0.1,1.2em)$) -|  (lig2.north);
    
    \node[block,
          right=0.7em of lig2,
          draw=blue!50!black,
          fill=blue!10!white
    ] (smi2) {SMILES};
    \draw[arr] (lig2) -- (smi2);
    \node[block,
          right=0.7em of smi2,
          draw=green!50!black,
          fill=green!10!white
    ] (lxyz2) {\texttt{<XYZ>}};
    \draw[arr] (smi2) -- (lxyz2);
    \node[block,
          right=1em of lxyz2,
          draw=blue!50!black,
          fill=blue!10!white
    ] (lxyzd2) {3D ligand \\ tokens};
    \draw[arr] (lxyz2) -- (lxyzd2);

    \node[below left=1.5em and 0em of lig2,anchor=west] (task2) {Pocket and Reward conditioned finetuning};
    \draw[-] (current bounding box.south east) -- (current bounding box.south west);

\end{tikzpicture}
\end{center}
\vspace{-0.5cm}
\caption{Data layout during the finetuning. Arrows show the tokens sequence order. Nodes such as \texttt{<POCKET>}
 show special tokens. Concrete examples are shown in Appendix \ref{appe:tokenization}.}
\vspace{-0.2cm}
\label{fig:ft_layout}
\end{wrapfigure}

As a result of the pretraining, BindGPT gains an understanding of a broad chemical space. This comprehensive understanding enables us to efficiently narrow it down through the supervised fine-tuning on a specialized dataset. During the supervised fine-tuning phase, we continue model training on CrossDocked 2020\citep{Francoeur2020}, which is a high-quality dataset containing aligned pocket-ligand pairs. Most of the prior methods subsample less than $1\%$ of the best pocket-ligand pairs and they don't benefit from it's diversity and scale. To obtain a bigger version of CrossDocked, we extract all intermediate ligand poses (with respect to the docking process), including the lower quality ones. Despite quite large size, CrossDocked was created by docking $14$k unique molecules into $3$k pockets \citep{Francoeur2020}. 
This is why we observed an dramatic overfitting when training on the $1\%$ version of CrossDocked and even on the full one. To alleviate that, we resort to two standard augmentation techniques used in drug discovery. First, we employ SMILES randomization \citep{smiles_rand}, which can heavily randomize one molecule by yielding 100-1000 different SMILES strings (all corresponding to that molecule). 
Second, we randomly rotate the 3D coordinates of the protein pocket and of the ligand (with the same rotation matrix). This way our model learns to \textit{understand structural and spatial properties of molecular binding beyond just token sequences}. 

Since the pretrained BindGPT is trained on both ligands (starting from the \texttt{<LIGAND>} token) and pockets (starting from the \texttt{<POCKET>} token), the information about the structure of both is learned by the model. 
In our finetuning setup, we represent each pocket-ligand pair as a string starting with the pocket string representation followed by the string representation of the ligand (See Section \ref{s:arm} for their description). 
Therefore, having learned them separately during pretraining, the finetuning exploits the independent knowledge of both pockets and ligands to learn a conditional dependency between them. In addition to that, since our version of CrossDocked contains both high and low score conformations, we test another version of context where we condition on the pocket and binding energy score obtained from the CrossDocked dataset (which is originally computed through the docking software \citep{vina1,vina2}). This way we can perform a variant of contrastive learning by learning the structure of good and bad examples. During evaluation of the model, we can sample molecules conditioned on some desired value of the binding affinity. The input layout for both versions is shown in Figure \ref{fig:ft_layout}.

\subsubsection{Reinforcement Learning}
\label{sec:rl_ft}

Despite the ubiquitous use of Reinforcement Learning (RL) for language models in Drug Discovery (see Section \ref{background} and Appendix \ref{app:efficient_training}), we did not find it been used within the pre-training paradigm of modern LLMs \citep{hoffmann2022training,kaplan2020scaling,instruct_gpt}.  Our main motivation to use RL after the pretraining/finetuning stages is to use the knowledge distilled into the model from massive amounts of less structured data. \textbf{We believe, this is the first work performing reinforcement learning on molecules that utilizes knowledge from pretraining and supervised finetuning.} Despite there are dozens of works doing RL with LMs on molecules, none of them do that within the LLM paradigm and none of them consider target-conditioned RL problem. In our opinion, the latter is primarily due to pocket-conditioned generation is not possible without large-scale pretraining as we show in the experimental section.

We apply the REINFORCE algorithm \citep{reinforce} for further model finetuning. It allows using the feedback (called \textit{reward}) from an external oracle to train model to generate even better structures compared to the ones it generates after the SFT stage. The resulting RL-finetuned model can generalize model and produce high affinity molecules even for the new pockets. 
In our procedure, on each training step we generate 3D structure of ligands for a batch of random protein pockets. Then we compute the reward using an external docking software that estimates the binding energy between the pocket and the generated ligand. The final step involves updating the language model with the batch of prompts (pockets), responses (ligands), and rewards (binding energies). We initially tested PPO \citep{schulman2017proximal} and REINVENT \citep{olivecrona2017molecular}, but found REINFORCE to be more stable for our project, which aligns with another recent finding in the field of RL applied to language models in NLP \citep{ahmadian2024basics}. Also, it's important to mention, that we apply the KL-penalty between the model's initialized and current state to stabilize the procedure. Further details such as hyperparameters can be found in the Appendix \ref{app:hps}.

\section{Results}

\begin{figure}[t]
    \vspace{-0.4cm}
    \centering
    \subfigure{
        \label{fig:conf_gen_samples}
        \includegraphics[width=0.4\linewidth]{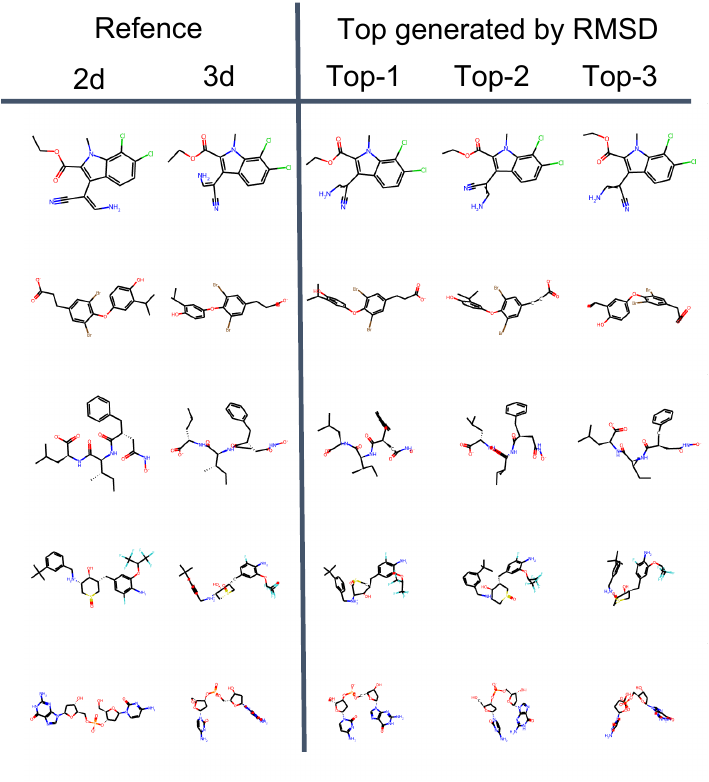}
    }
    \subfigure{
        \label{fig:conf_gen}
        \includegraphics[width=0.45\linewidth]{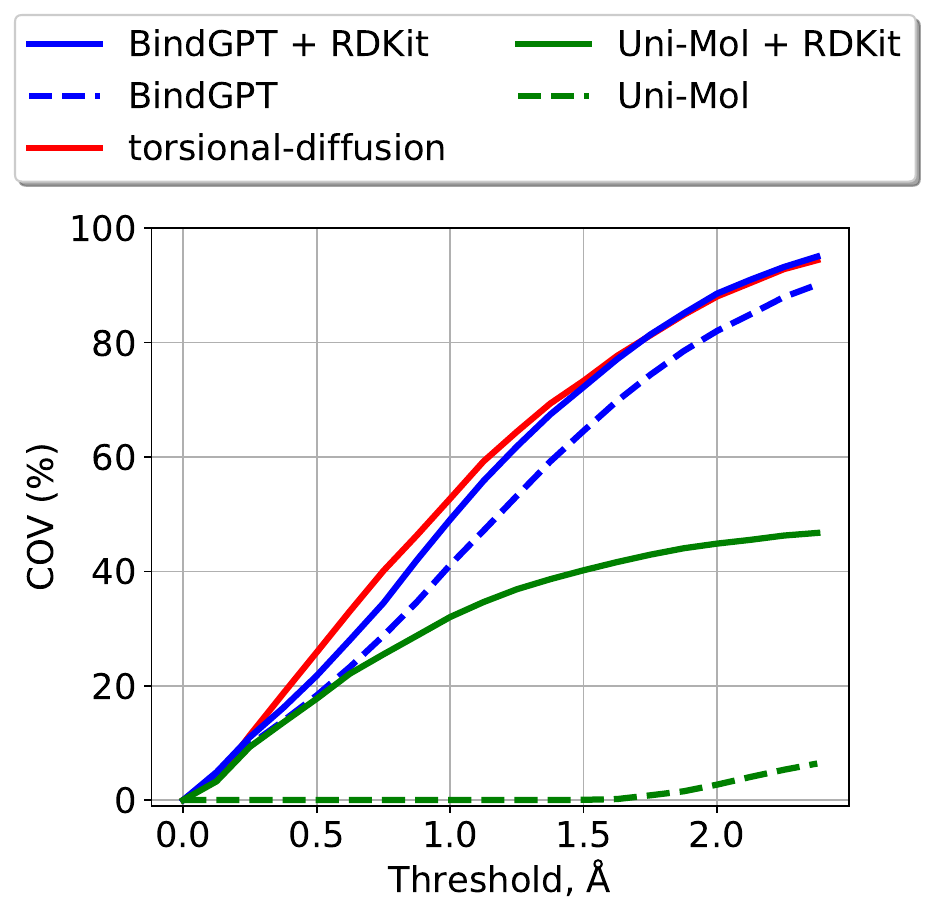}
    }
    \caption{{\bf (left)} Sampled conformations for reference molecules from the Platinum dataset (See Appendix \ref{app:more_samples} for a higher resolution image) {\bf (right)} RMSD Coverage metric calculated on the Platinum dataset for the 3D conformation generation task.}
    \vspace{-0.5cm}
\end{figure}

In this section, we describe our experimental results. We start with a brief data description followed by the description of the three 3D molecular generative tasks: the 3D generative modeling of molecules and conformation generation given molecular graph (in Section \ref{sec:gen_3d}) and then pocket-conditioned generation (in Section \ref{sec:pock_cond}).

For pretraining, we use a large 3D molecular dataset proposed by the authors of the Uni-Mol model \citep{zhou2023unimol}. The dataset contains 208M conformations for 12M molecules and 3.2M spatial structures of protein pockets. For finetuning in the pocket-conditioned generation task, we use the aforementioned CrossDocked dataset which contains aligned pocket-molecule pairs. Our filtration of the dataset has around 27M pocket-ligand pairs covering a cross product of 14k molecules with 3k pockets (not all of the pairs are present, and for some of them, there is more than one pose, each with different score). We also hold out a set of 100 pockets from the training data for evaluating the model performance. For the tasks of 3D molecule and 3D conformer generation (Section \ref{sec:gen_3d}), to make comparisons with baselines more fair, we also finetune the model on the GEOM-DRUGS \citep{geom_drugs} dataset, with drug-like molecules having high-quality 3D molecular conformations. This dataset contains 27M conformations for 300k molecules, and it serves as a standard benchmark for the machine learning-based 3D molecular generators. Finally, we use the Platinum \citep{Platinum} dataset as a hold-out evaluation dataset to test our model and baselines on zero-shot conformer generation. Platinum dataset contains the best-in-class experimentally validated conformations for testing conformer generation software.

\subsection{Generative Modeling of 3D molecules}
\label{sec:gen_3d}

\begin{table}[h!]
    \centering
    \begin{adjustbox}{width=0.95\textwidth}
    {\setlength{\extrarowheight}{5pt}
\begin{tabular}{l|cccccc}
 method & Valid ($\uparrow$) & SA ($\uparrow$)& QED ($\uparrow$) & Lipinski ($\uparrow$) & RMSD ($\downarrow$) & Time,s ($\downarrow$) \\
 \hline
 XYZ-TF  & $12.87\%$ & $0.21$ & $0.30$ & $4.79$ & - & 165 \\
 BindGPT (Ours) & $\mathbf{98.58\%}$ & $\mathbf{0.77}$ & $\mathbf{0.59}$ & $\mathbf{4.86}$ & $\mathbf{0.89}$ & $\mathbf{13}$ \\ 
\cline{1-7}
XYZ-TF (\texttt{H})  & $17.86\%$ & $0.54$ & $0.37$ & $4.82$ & - & $394$ \\
BindGPT (\texttt{H}) (Ours) & $\mathbf{77.33\%}$ & $\mathbf{0.78}$ & $\mathbf{0.61}$ & $\mathbf{4.91}$ & $\mathbf{3.44}$ & $\mathbf{156}$ \\ 
\end{tabular}}
\end{adjustbox}
    \caption{Generative metrics for the molecule generation task after the pretraining. 
    \texttt{(H)} is explicit hydrogens are generated with molecules. For XYZ-TF, the RMSD calculation algorithm failed to converge.}
    \label{tab:3d_gen}
\end{table}

\begin{table}[t!]
\vspace{-0.4cm}
\begin{center}
\begin{adjustbox}{width=0.99\textwidth}
\begin{tabular}{ c | c | c | c | c} 
 Group & Metrics & EDM & MolDiff & \makecell{BindGPT \\ (Ours)} \\ 
 \hline
 \multirow{3}{*}{Druglikeness} & QED ($\uparrow$) & 0.558 & \textbf{0.668} & 0.616 \\
  & SA ($\uparrow$) & 0.568 & \textbf{0.874} & 0.826 \\
  & Lipinski ($\uparrow$) & 4.923 & \textbf{4.986} & 4.896 \\
 \hline
\multirow{3}{*}{3D structures} & JS. bond lengths ($\downarrow$) & 0.246 & 0.365 & \textbf{0.029} \\
 & JS. bond angles ($\downarrow$) & 0.282 & 0.155 & \textbf{0.075} \\
 & JS. dihedral angles ($\downarrow$) & 0.328 & 0.162 & \textbf{0.098}  \\
 \hline
\multirow{4}{*}{Bonds} & JS. num. bonds per atoms ($\downarrow$) & 0.139 & \textbf{0.115} & 0.160 \\
& JS. freq. bond types  ($\downarrow$) & 0.378 & 0.163 & \textbf{0.045}  \\
& JS. freq. bond pairs  ($\downarrow$) & 0.396 & 0.136 &  \textbf{0.043} \\
& JS. freq. bond triplets  ($\downarrow$) & 0.449 & 0.125 & \textbf{0.042} \\
\hline
\multirow{3}{*}{Rings} & JS. num. rings ($\downarrow$) & 0.106 & \textbf{0.062} & 0.094 \\
 & JS. num. n-sized rings ($\downarrow$) & 0.107 & 0.092 & \textbf{0.023} \\
 & Num. Intersecting rings ($\uparrow$) & 3.667 & 8.000 & \textbf{9.000} \\
\hline
& Time for 1000 valid molecules, s ($\downarrow$) & $1.4\times 10^6$ & $7500$ & $\mathbf{200}$ \\
\end{tabular}
\end{adjustbox}
\end{center}
\caption{The qualities of the generated 3D molecules after finetuning on GEOM-DRUGS.}
\vspace{-0.5cm}
\label{tab:unconditional_generation}
\end{table}

\textbf{Metrics.}  We provide the validity ($\uparrow$) of generated molecules and druglikeness metrics - SA ($\uparrow$), QED ($\uparrow$), and Lipinski ($\uparrow$) that are agnostic to 3D but measure how likely the molecule to be a drug. Also, we adopt a range of distribution metrics that were used for the MolDiff method \citep{moldiff}. Those metrics measure the discrepancy between true and modelled molecular distributions by computing the Jensen–Shannon divergences on the set of molecular properties and features distributions. We compute RMSD (Root-Mean-Squared-Distance) ($\downarrow$) - which measures the quality of 3D structures by aligning the generated one with the one from RDkit (i.e., we regenerate conformer via RDkit) and computing the atomwise distance. 
Finally, we measure the time needed to generate 1K valid 3D molecules on one GPU. Note that this choice of metrics is standard for this task (see \citet{moldiff} for a more detailed description of them). For the 3D conformation generation given molecule task, we compute the RMSD-coverage ($\uparrow$) metric. This is a standard performance metric for 3D conformer generation models (see e.g. \citet{jing2022torsional}). It is represented by the cumulative distribution function of RMSD between generated and reference conformers. The metric is a function of the threshold $x$: $P(\text{RMSD} < x)$. An ideal model should have as high metric value as possible for as low thresholds as possible.

\textbf{Baselines.} For the molecule generation task, we consider the current best 3D generative models. EDM \citep{hoogeboom2022edm} and MolDiff \citep{moldiff} are task-specialized diffusion models for 3D molecule generation. XYZ-Transformer \citep{flamshepherd2023language} is another 3D molecular transformer that was proposed for small-scale data. Note that XYZ-TF is the only model capable of large scale pretraining besides our model, so we pretrain only XYZ-TF and BindGPT on the Uni-Mol data. We also do the GEOM-DRUGS evaluation, where we report MolDiff and EDM trained on the full dataset and for BindGPT finetuned on the same version of it. For conformer generation, we compare BindGPT with the current state-of-the-art methods, Torsional Diffusion \citep{jing2022torsional} and the Uni-Mol model \citep{zhou2023unimol}. The former is a specialized SE(3)-equivariant diffusion model capable of conformation generation only. The latter is a modified BERT \citep{devlin2019bert}. As a coordinate-level encoder LM, the Uni-Mol model needs input coordinates to generate a conformation, which is why this model uses RDKit as a tool for initializing coordinates.

\textbf{Results.} The molecular generative modeling results are shown in Tables \ref{tab:unconditional_generation} and \ref{tab:3d_gen}. First, the pretrained BindGPT model consistently outperforms the XYZ-TF baseline both without and with explicit hydrogens. The latter is a much more challenging task and almost no baseline methods can do that (except EDM, which is not scalable) since reconstructing hydrogens can be done on a post-processing step but explicit modeling of them makes the molecule size several times larger. BindGPT is the first model capable of modeling hydrogen explicitly at such large scale. Also, XYZ-TF has a very low validity rate due to the need of graph reconstruction. 
Next, for the methods trained on the GEOM-DRUGS dataset, BindGPT (being finetuned on this data) shows state-of-the-art performance scores for nearly all distributional evaluation metrics. Even though BindGPT does not outperform MolDiff in Druglikeness, that could be explained by a smaller vocabulary of the \citep{moldiff},  containing only frequent atoms. 
For the conformation generation task, the current best baseline is Torsional Diffusion (TD) \citep{jing2022torsional}. We use the Platinum dataset to compare TD trained on GEOM-DRUGS with Uni-Mol-BERT and BindGPT, both of which are pretrained and finetuned on the same data. Figure \ref{fig:conf_gen} shows the results for zero-shot evaluations on Platinum. Surprisingly, Uni-Mol fails to generalize to this new dataset (even assisted by RDKit), which we think is because of its structural diversity. BindGPT, in contrast, is capable of matching the performance of TD when assisted by the RDKit tool and having a small gap when not. All the above results demonstrate the generalisability of our model - none of the baselines is able to solve this wide range of task at this level of quality.

\subsection{Pocket-conditioned Molecule Generation}
\label{sec:pock_cond}

\begin{figure}[t]
    \vspace{-0.8cm}
    \centering
    \label{fig:pocket_conditioned}
    \includegraphics[width=0.7\linewidth]{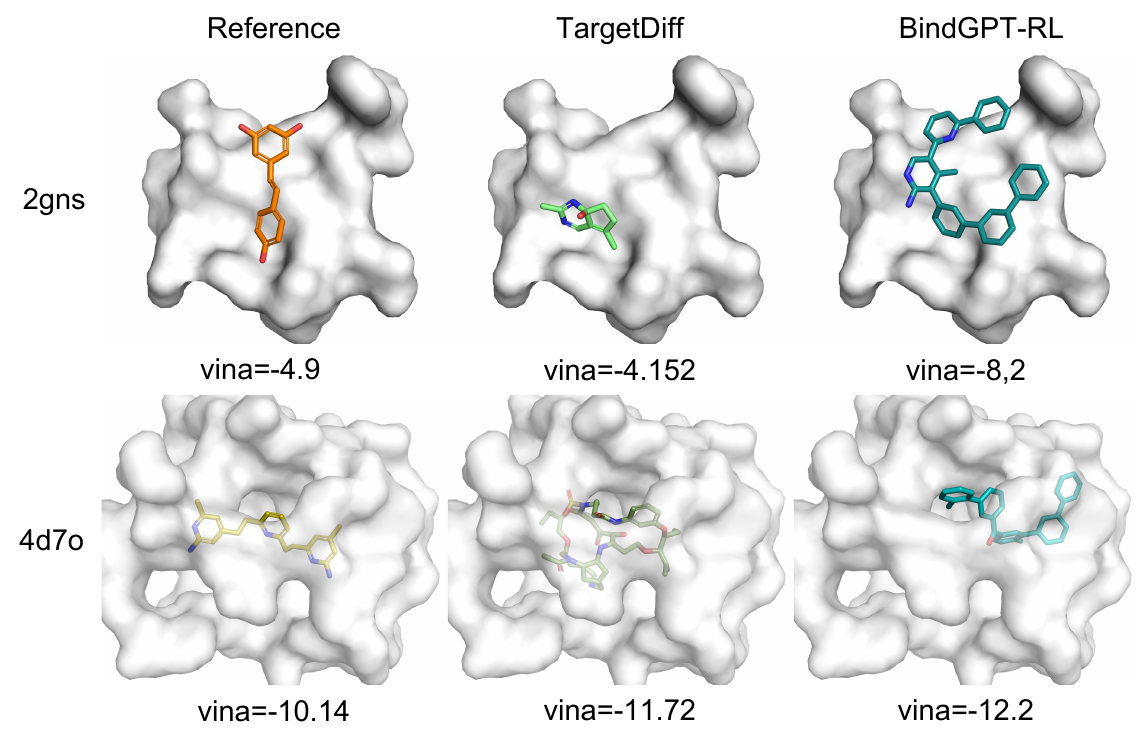}
    \caption{Examples of binding poses and Vina scores ($\downarrow$) for 2gns and 4d7o pockets.}
    \vspace{-0.5cm}
\end{figure}

\textbf{Metrics.} The main metrics for this task include the measure of ligand-pocket affinity; and druglikeness of the ligand. The first one is represented via the binding energy ($\downarrow$) computed by the QVINA \citep{qvina}  docking software, while the second one comprises the aforementioned druglikeness metrics (SA ($\uparrow$), QED ($\uparrow$), and Lipinsky ($\uparrow$)). For each baseline we report the time required to generate $100$ valid molecules for one pocket.

\textbf{Baselines.} Apart from the BindGPT model, we include the baselines such as 3D diffusion model (TargetDiff \citep{guan3d}) and autoregressive Graph Neural Network (Pocket2Mol \citep{peng2022pocket2mol}).
Note that none of the baselines perform large-scale pretraining. Instead, they resort to heavy inductive biases to efficiently learn from small-scale data.

\textbf{Results.} Performance of our approach is summarized in Table \ref{tab:pocket_cond_gen}. We depict the performance of three version of BindGPT. First, \texttt{BindGPT-FT} is a model finetuned on the complete CrossDocked data (the data layout as described in Figure \ref{fig:ft_layout}, top), i.e. both good and bad binding pairs. This model serves as an initialization for the reinforcement learning model. Second, \texttt{BindGPT-RFT} is the model finetuned on CrossDocked with the reward in the context. To get higher affinity molecules from that model, we condition the model on random binding energy values within $[-12, -10]$, which are the best scores observed by the model (in around $0.1\%$ of examples). Finally, the \texttt{BindGPT-RL} model is trained with RL (see Section \ref{sec:rl_ft} and Appendix \ref{app:rl_desc}) from the \texttt{BindGPT-FT} initialization. Our main conclusion is that the RL finetuned model can learn to search the space of binding molecules much more efficiently and significantly outperforms all the previous best baselines in terms of the binding energy.

\begin{table}[h!]
    \centering
    \begin{adjustbox}{width=0.95\textwidth}
    {\setlength{\extrarowheight}{5pt}
\begin{tabular}{l|ccccc}
Method & Vina score ($\downarrow$) & SA ($\uparrow$)& QED ($\uparrow$) & Lipinski ($\uparrow$) \\
 \hline
 Pocket2Mol & $-7.15 \pm 4.89$ & $0.75 \pm 0.12$ & $\mathbold{0.57 \pm 0.15}$ & $\mathbf{4.88 \pm 0.37}$ \\
 TargetDiff & $-7.80 \pm 3.61$ & $0.58\pm 0.12$ & $0.48 \pm 0.19$ & $4.51\pm 0.85$ \\
 BindGPT-FT (Ours) & $-5.44 \pm 2.09$ & $0.78\pm 0.10$ & $0.50 \pm 0.17$ & $4.72 \pm 0.70$ \\
 BindGPT-RFT (Ours) &  $-7.24 \pm 1.68$ & $0.74 \pm 0.11$  & $0.48 \pm 0.22$ & $4.32  \pm 1.25$  \\
 BindGPT-RL (Ours) & $\mathbf{-8.60\pm 1.90}$ & $\mathbf{0.84\pm 0.05}$ & $0.43\pm 0.17$ & $4.81\pm 0.52$  \\
 
\end{tabular}}
\end{adjustbox}
    \caption{Generative metrics for the pocket-conditioned generation task.
    }
    \label{tab:pocket_cond_gen}
\end{table}

\vspace{-0.4cm}
\section{Discussion and Conclusion}

In this work, we presented BindGPT, a scalable framework for training capable language models that can generate 3D molecules as text. Through a series of studies on a range of different 3D molecular generative tasks, we demonstrate the generality of our approach as it can solve each of them by matching or surpassing the baselines. Notably, our method does not have any inductive biases about the generative domain acting as a general and data-driven approach. Unlike all the baselines which have strong inductive biases, our method solves each downstream task without any such assumptions. The task of a particular interest in our work is the pocket-based molecule generation where our model outperforms all the baselines with a large margin. We show that the large-scale pretraining paradigm can be efficiently transfered from NLP to the 3D drug discovery.

\subsubsection*{Acknowledgments}

 Sarath Chandar is supported by the Canada CIFAR AI Chairs program, the Canada Research Chair in Lifelong Machine Learning, and the NSERC Discovery Grant.

\bibliography{colm2024_conference}
\bibliographystyle{colm2024_conference}

\newpage
\appendix
\section{Tokenization and Data Representation}
\label{appe:tokenization}
\begin{figure}[!h]
    \centering
    \begin{tikzpicture}[scale=1]
    \node[] (root) {\texttt{<LIGAND>[H]c1c(F)c([H])c2c(C(F)(F)F)c([H])c(C\#N)nc2c1[H]}};
    \node[below left= 0.5em and 0em of root, anchor=west] (xyz) {\texttt{<XYZ>}};
    \node[below left = 0.5em and 0em of xyz, anchor=west] (num) {\texttt{21}};
    \node[below left = 0.5em and 0em of num, anchor=west] (c1) {\texttt{\phantom{-}2.775 -0.640 \phantom{-}2.950}};
    \node[below left = 0.5em and 0em of c1, anchor=west] (c2) {\texttt{\phantom{-}2.078 -0.379 \phantom{-}2.160}};
    \node[below left = 0.5em and 0em of c2, anchor=west] (c3) {\texttt{\phantom{-}2.123 \phantom{-}0.876 \phantom{-}1.583}};
    \node[below left = 0.5em and 0em of c3, anchor=west] (c4) {\texttt{\phantom{-}3.028 \phantom{-}1.767 \phantom{-}2.004}};
    \node[below left = 0.5em and 0em of c4, anchor=west] (c5) {\texttt{\phantom{-}1.236 \phantom{-}1.224 \phantom{-}0.571}};
    \node[below left = 0.5em and 0em of c5, anchor=west] (c6) {\texttt{\phantom{-}1.321 \phantom{-}2.226 \phantom{-}0.157}};
    \node[below left = 0.5em and 0em of c6, anchor=west] (c7) {\texttt{\phantom{-}0.270 \phantom{-}0.300 \phantom{-}0.109}};
    \node[below left = 0.5em and 0em of c7, anchor=west] (c8) {\texttt{-0.667 \phantom{-}0.577 -0.917}};
    \node[below left = 0.5em and 0em of c8, anchor=west] (c9) {\texttt{-0.709 \phantom{-}1.913 -1.616}};
    \node[below left = 0.5em and 0em of c9, anchor=west] (c10) {\texttt{-0.953 \phantom{-}2.943 -0.761}};
    \node[below left = 0.5em and 0em of c10, anchor=west] (c11) {\texttt{\phantom{-}0.457 \phantom{-}2.211 -2.248}};
    \node[below left = 0.5em and 0em of c11, anchor=west] (c12) {\texttt{-1.677 \phantom{-}1.990 -2.572}};
    \node[below left = 0.5em and 0em of c12, anchor=west] (c13) {\texttt{-1.580 -0.419 -1.293}};
    \node[below left = 0.5em and 0em  of c13, anchor=west] (c14) {\texttt{-2.313 -0.241 -2.077}};
    \node[below left = 0.5em and 0em of c14, anchor=west] (c15) {\texttt{-1.534 -1.641 -0.648}};
    \node[below left = 0.5em and 0em of c15, anchor=west] (c16) {\texttt{-2.449 -2.694 -0.998}};
    \node[below left = 0.5em and 0em of c16, anchor=west] (c17) {\texttt{-3.207 -3.516 -1.313}};
    \node[below left = 0.5em and 0em of c17, anchor=west] (c18) {\texttt{-0.650 -1.928 \phantom{-}0.332}};
    \node[below left = 0.5em and 0em of c18, anchor=west] (c19) {\texttt{\phantom{-}0.227 -0.985 \phantom{-}0.701}};
    \node[below left = 0.5em and 0em of c19, anchor=west] (c20) {\texttt{\phantom{-}1.129 -1.298 \phantom{-}1.711}};
    \node[below left = 0.5em and 0em of c20, anchor=west] (c21) {\texttt{\phantom{-}1.093 -2.287 \phantom{-}2.164}};
    \end{tikzpicture}
    \caption{An example of a 3D molecule encoded as text that the model trains to predict. The sequence starts with a special token indicating the beginning of a small molecule: \texttt{<LIGAND>} followed by a SMILES string, tokenized character-level. Next follows the \texttt{<XYZ>} special token marking the end of SMILES and the beginning of the coordinate part of the output. The coordinate part of the output starts with the number of atoms in the molecule followed by a series of 3D coordinates for each atom. Unlike the standard XYZ format, our representation doesn't include atom types before each coordinate due to the preceding SMILES string, which already provides atom sequence and connectivity. In our case, the order of 3D coordinates correspond to the order of atoms as they appear in the SMILES string. Each coordinate triplet is encoded with six tokens: one token specifies the integer part with a sign and another one defines a fractional part of the floating point number.}
    \label{fig:example_tokens}
\end{figure}

\newpage
\begin{figure}[!h]
    \centering
    \begin{tikzpicture}[scale=1]
    \node[] (root) {\texttt{\sethlcolor{lightblue}\hl{<LIGAND>}\sethlcolor{lightgreen}\hl{[}\sethlcolor{lightpink}\hl{H}\sethlcolor{lightyellow}\hl{]}\sethlcolor{lightblue}\hl{c}\sethlcolor{darkgreen}\hl{1}\sethlcolor{darkblue}\hl{c}\sethlcolor{darkred}\hl{(}\sethlcolor{lightgreen}\hl{F}\sethlcolor{lightpink}\hl{)}\sethlcolor{lightyellow}\hl{c}\sethlcolor{lightblue}\hl{(}\sethlcolor{darkgreen}\hl{[}\sethlcolor{darkblue}\hl{H}\sethlcolor{darkred}\hl{]}\sethlcolor{lightgreen}\hl{)}\sethlcolor{lightpink}\hl{c}\sethlcolor{lightyellow}\hl{2}\sethlcolor{lightblue}\hl{c}\sethlcolor{darkgreen}\hl{(}\sethlcolor{darkblue}\hl{C}\sethlcolor{darkred}\hl{(}\sethlcolor{lightgreen}\hl{F}\sethlcolor{lightpink}\hl{)}\sethlcolor{lightyellow}\hl{(}\sethlcolor{lightblue}\hl{F}\sethlcolor{darkgreen}\hl{)}\sethlcolor{darkblue}\hl{F}\sethlcolor{darkred}\hl{)}\sethlcolor{lightgreen}\hl{c}\sethlcolor{lightpink}\hl{(}\sethlcolor{lightyellow}\hl{[}\sethlcolor{lightblue}\hl{H}\sethlcolor{darkgreen}\hl{]}\sethlcolor{darkblue}\hl{)}\sethlcolor{darkred}\hl{c}\sethlcolor{lightgreen}\hl{(}\sethlcolor{lightpink}\hl{C}\sethlcolor{lightyellow}\hl{\#}\sethlcolor{lightblue}\hl{N}\sethlcolor{darkgreen}\hl{)}\sethlcolor{darkblue}\hl{n}\sethlcolor{darkred}\hl{c}\sethlcolor{lightgreen}\hl{2}\sethlcolor{lightpink}\hl{c}\sethlcolor{lightyellow}\hl{1}\sethlcolor{lightblue}\hl{[}\sethlcolor{darkgreen}\hl{H}\sethlcolor{darkblue}\hl{]}}};
    \node[below left= 0.5em and 0em of root, anchor=west] (xyz) {\texttt{\sethlcolor{darkred}\hl{<XYZ>}}};
    \node[below left = 0.5em and 0em of xyz, anchor=west] (num) {\texttt{\sethlcolor{lightgreen}\hl{21}}};
    \node[below left = 0.5em and 0em of num, anchor=west] (c1) {\texttt{\phantom{-}\sethlcolor{lightgreen}\hl{2}\sethlcolor{lightpink}\hl{.775} \sethlcolor{lightyellow}\hl{-0}\sethlcolor{lightblue}\hl{.640} \phantom{-}\sethlcolor{darkgreen}\hl{2}\sethlcolor{darkblue}\hl{.950}}};
    \node[below left = 0.5em and 0em of c1, anchor=west] (c2) {\texttt{\phantom{-}\sethlcolor{darkred}\hl{2}\sethlcolor{lightgreen}\hl{.078} \sethlcolor{lightpink}\hl{-0}\sethlcolor{lightyellow}\hl{.379} \phantom{-}\sethlcolor{lightblue}\hl{2}\sethlcolor{darkgreen}\hl{.160}}};
    \node[below left = 0.5em and 0em of c2, anchor=west] (c3) {\texttt{\phantom{-}\sethlcolor{darkblue}\hl{2}\sethlcolor{darkred}\hl{.123} \phantom{-}\sethlcolor{lightgreen}\hl{0}\sethlcolor{lightpink}\hl{.876} \phantom{-}\sethlcolor{lightyellow}\hl{1}\sethlcolor{lightblue}\hl{.583}}};
    \node[below left = 0.5em and 0em of c3, anchor=west] (c4) {\texttt{\phantom{-}\sethlcolor{darkgreen}\hl{3}\sethlcolor{darkblue}\hl{.028} \phantom{-}\sethlcolor{darkred}\hl{1}\sethlcolor{lightgreen}\hl{.767} \phantom{-}\sethlcolor{lightpink}\hl{2}\sethlcolor{lightyellow}\hl{.004}}};
    \node[below left = 0.5em and 0em of c4, anchor=west] (c5) {\texttt{\phantom{-}\sethlcolor{lightblue}\hl{1}\sethlcolor{darkgreen}\hl{.236} \phantom{-}\sethlcolor{darkblue}\hl{1}\sethlcolor{darkred}\hl{.224} \phantom{-}\sethlcolor{lightgreen}\hl{0}\sethlcolor{lightpink}\hl{.571}}};
    \node[below left = 0.5em and 0em of c5, anchor=west] (c6) {\texttt{\phantom{-}\sethlcolor{lightyellow}\hl{1}\sethlcolor{lightblue}\hl{.321} \phantom{-}\sethlcolor{darkgreen}\hl{2}\sethlcolor{darkblue}\hl{.226} \phantom{-}\sethlcolor{darkred}\hl{0}\sethlcolor{lightgreen}\hl{.157}}};
    \node[below left = 0.5em and 0em of c6, anchor=west] (c7) {\texttt{\phantom{-}\sethlcolor{lightpink}\hl{0}\sethlcolor{lightyellow}\hl{.270} \phantom{-}\sethlcolor{lightblue}\hl{0}\sethlcolor{darkgreen}\hl{.300} \phantom{-}\sethlcolor{darkblue}\hl{0}\sethlcolor{darkred}\hl{.109}}};
    \node[below left = 0.5em and 0em of c7, anchor=west] (c8) {\texttt{\sethlcolor{lightgreen}\hl{-0}\sethlcolor{lightpink}\hl{.667} \phantom{-}\sethlcolor{lightyellow}\hl{0}\sethlcolor{lightblue}\hl{.577} \sethlcolor{darkgreen}\hl{-0}\sethlcolor{darkblue}\hl{.917}}};
    \node[below left = 0.5em and 0em of c8, anchor=west] (c9) {\texttt{\sethlcolor{darkred}\hl{-0}\sethlcolor{lightgreen}\hl{.709} \phantom{-}\sethlcolor{lightpink}\hl{1}\sethlcolor{lightyellow}\hl{.913} \sethlcolor{lightblue}\hl{-1}\sethlcolor{darkgreen}\hl{.616}}};
    \node[below left = 0.5em and 0em of c9, anchor=west] (c10) {\texttt{\sethlcolor{darkblue}\hl{-0}\sethlcolor{darkred}\hl{.953} \phantom{-}\sethlcolor{lightgreen}\hl{2}\sethlcolor{lightpink}\hl{.943} \sethlcolor{lightyellow}\hl{-0}\sethlcolor{lightblue}\hl{.761}}};
    \node[below left = 0.5em and 0em of c10, anchor=west] (c11) {\texttt{\phantom{-}\sethlcolor{darkgreen}\hl{0}\sethlcolor{darkblue}\hl{.457} \phantom{-}\sethlcolor{darkred}\hl{2}\sethlcolor{lightgreen}\hl{.211} \sethlcolor{lightpink}\hl{-2}\sethlcolor{lightyellow}\hl{.248}}};
    \node[below left = 0.5em and 0em of c11, anchor=west] (c12) {\texttt{\sethlcolor{lightblue}\hl{-1}\sethlcolor{darkgreen}\hl{.677} \phantom{-}\sethlcolor{darkblue}\hl{1}\sethlcolor{darkred}\hl{.990} \sethlcolor{lightgreen}\hl{-2}\sethlcolor{lightpink}\hl{.572}}};
    \node[below left = 0.5em and 0em of c12, anchor=west] (c13) {\texttt{\sethlcolor{lightyellow}\hl{-1}\sethlcolor{lightblue}\hl{.580} \sethlcolor{darkgreen}\hl{-0}\sethlcolor{darkblue}\hl{.419} \sethlcolor{darkred}\hl{-1}\sethlcolor{lightgreen}\hl{.293}}};
    \node[below left = 0.5em and 0em  of c13, anchor=west] (c14) {\texttt{\sethlcolor{lightpink}\hl{-2}\sethlcolor{lightyellow}\hl{.313} \sethlcolor{lightblue}\hl{-0}\sethlcolor{darkgreen}\hl{.241} \sethlcolor{darkblue}\hl{-2}\sethlcolor{darkred}\hl{.077}}};
    \node[below left = 0.5em and 0em of c14, anchor=west] (c15) {\texttt{\sethlcolor{lightgreen}\hl{-1}\sethlcolor{lightpink}\hl{.534} \sethlcolor{lightyellow}\hl{-1}\sethlcolor{lightblue}\hl{.641} \sethlcolor{darkgreen}\hl{-0}\sethlcolor{darkblue}\hl{.648}}};
    \node[below left = 0.5em and 0em of c15, anchor=west] (c16) {\texttt{\sethlcolor{darkred}\hl{-2}\sethlcolor{lightgreen}\hl{.449} \sethlcolor{lightpink}\hl{-2}\sethlcolor{lightyellow}\hl{.694} \sethlcolor{lightblue}\hl{-0}\sethlcolor{darkgreen}\hl{.998}}};
    \node[below left = 0.5em and 0em of c16, anchor=west] (c17) {\texttt{\sethlcolor{darkblue}\hl{-3}\sethlcolor{darkred}\hl{.207} \sethlcolor{lightgreen}\hl{-3}\sethlcolor{lightpink}\hl{.516} \sethlcolor{lightyellow}\hl{-1}\sethlcolor{lightblue}\hl{.313}}};
    \node[below left = 0.5em and 0em of c17, anchor=west] (c18) {\texttt{\sethlcolor{darkgreen}\hl{-0}\sethlcolor{darkblue}\hl{.650} \sethlcolor{darkred}\hl{-1}\sethlcolor{lightgreen}\hl{.928} \phantom{-}\sethlcolor{lightpink}\hl{0}\sethlcolor{lightyellow}\hl{.332}}};
    \node[below left = 0.5em and 0em of c18, anchor=west] (c19) {\texttt{\phantom{-}\sethlcolor{lightblue}\hl{0}\sethlcolor{darkgreen}\hl{.227} \sethlcolor{darkblue}\hl{-0}\sethlcolor{darkred}\hl{.985} \phantom{-}\sethlcolor{lightgreen}\hl{0}\sethlcolor{lightpink}\hl{.701}
}};
    \node[below left = 0.5em and 0em of c19, anchor=west] (c20) {\texttt{\phantom{-}\sethlcolor{lightyellow}\hl{1}\sethlcolor{lightblue}\hl{.129} \sethlcolor{darkgreen}\hl{-1}\sethlcolor{darkblue}\hl{.298} \phantom{-}\sethlcolor{darkred}\hl{1}\sethlcolor{lightgreen}\hl{.711}}};
    \node[below left = 0.5em and 0em of c20, anchor=west] (c21) {\texttt{\phantom{-}\sethlcolor{lightpink}\hl{1}\sethlcolor{lightyellow}\hl{.093} \sethlcolor{lightblue}\hl{-2}\sethlcolor{darkgreen}\hl{.287} \phantom{-}\sethlcolor{darkblue}\hl{2}\sethlcolor{darkred}\hl{.164}}};
    \end{tikzpicture}
    \caption{The visualization of the tokenization scheme for molecules in our language model. We use the same text as in Figure \ref{fig:example_tokens}. Text is colored according to how tokenization is performed. Note that untokenized symbols (e.g. white spaces and new line separators) are not passed to the model since they don't carry any useful information. Also note that the coloring is made only for visualization purposes.}
    \label{fig:example_tokens_vis}
\end{figure}
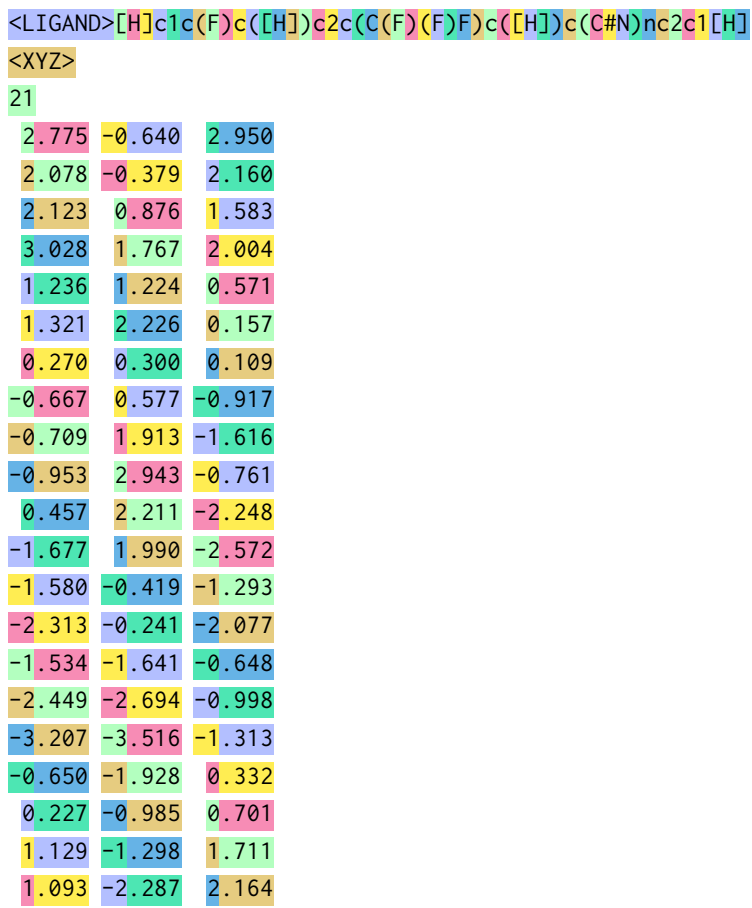

\newpage
\begin{figure}[!ht]
    \centering
    \begin{tikzpicture}[scale=1]
    \node[anchor=west] (root) {\texttt{<POCKET>NCACOCCCCCCCONCACOCCCCNCACOCCCNCACOCCNCCNNCACOCCCCNCACOCNCAC}};
    \node[below left= 0.5em and 0em of root, anchor=west] (root2) {\texttt{OCCCNCNNNCACOCCCNCACOCCONNCACOCCONNCACOCCOONCACOCCCCCCCNCACOCCSC}};
    \node[below left= 0.5em and 0em of root2, anchor=west] (xyz) {\texttt{<XYZ>}}; 
    \node[below left = 0.5em and 0em of xyz, anchor=west] (c1) {\texttt{-4.991 \phantom{-}4.794 \phantom{-}6.134}};
    \node[below left = 0.5em and 0em of c1, anchor=west] (c2) {\texttt{\phantom{-}3.067 \phantom{-}2.185 -5.773}};
    \node[below left = 0.5em and 0em of c2, anchor=west] (c3) {\texttt{\phantom{-}0.121 -1.334 \phantom{-}3.936}};
    \node[below left = 0.5em and 0em of c3, anchor=west] (c4) {\texttt{\phantom{-}7.077 \phantom{-}1.335 \phantom{-}2.009}};
    \node[below left = 0.5em and 0em of c4, anchor=west] (c5) {\texttt{\phantom{-}1.134 -7.460 -5.195}};
    \node[below left = 0.5em and 0em of c5, anchor=west] (c6) {\texttt{\phantom{-}2.384 -5.084 \phantom{-}0.318}};
    \node[below left = 0.5em and 0em of c6, anchor=west] (c7) {\texttt{-5.272 -7.393 -5.431}};
    \node[below left = 0.5em and 0em of c7, anchor=west] (c8) {\texttt{\phantom{-}7.391 -1.954 \phantom{-}0.092}};
    \node[below left = 0.5em and 0em of c8, anchor=west] (c9) {\texttt{-0.887 -4.613 \phantom{-}2.238}};
    \node[below left = 0.5em and 0em of c9, anchor=west] (c10) {\texttt{\phantom{-}0.488 \phantom{-}5.700 -2.473}};
    \node[below left = 0.5em and 0em of c10, anchor=west] (c11) {\texttt{-3.573 \phantom{-}8.085 \phantom{-}4.613}};
    \node[below left = 0.5em and 0em of c11, anchor=west] (c12) {\texttt{\phantom{-}3.905 -1.313 -4.534}};
    \node[below left = 0.5em and 0em of c12, anchor=west] (c13) {\texttt{-10.845 7.057 \phantom{-}4.070}};
    \end{tikzpicture}
    \caption{An example of the protein pocket encoded as text that the model trains to predict during pretraining. Sequence starts with a special token \texttt{<POCKET>} marking the beginning of the protein pocket sequence. After that, we write a sequence of heavy atoms ignoring the edge structure of the pocket part of the protein molecule. After that, we write 3D coordinates of Alpha-Carbon atoms, each of which appears only once per aminoacid. The tokenization of the 2D part of the pocket is character level (except for \texttt{CA}) and for the 3D part it is two tokens per real number like in the ligand tokenization.}
    \label{fig:example_pockets}
\end{figure}

\begin{figure}[!ht]
    \centering
    \begin{tikzpicture}[scale=1]
    \node[anchor=west] (root) {\texttt{\sethlcolor{lightblue}\hl{<POCKET>}\sethlcolor{lightgreen}\hl{N}\sethlcolor{lightpink}\hl{CA}\sethlcolor{lightyellow}\hl{C}\sethlcolor{lightblue}\hl{O}\sethlcolor{darkgreen}\hl{C}\sethlcolor{darkblue}\hl{C}\sethlcolor{darkred}\hl{C}\sethlcolor{lightgreen}\hl{C}\sethlcolor{lightpink}\hl{C}\sethlcolor{lightyellow}\hl{C}\sethlcolor{lightblue}\hl{C}\sethlcolor{darkgreen}\hl{O}\sethlcolor{darkblue}\hl{N}\sethlcolor{darkred}\hl{CA}\sethlcolor{lightgreen}\hl{C}\sethlcolor{lightpink}\hl{O}\sethlcolor{lightyellow}\hl{C}\sethlcolor{lightblue}\hl{C}\sethlcolor{darkgreen}\hl{C}\sethlcolor{darkblue}\hl{C}\sethlcolor{darkred}\hl{N}\sethlcolor{lightgreen}\hl{CA}\sethlcolor{lightpink}\hl{C}\sethlcolor{lightyellow}\hl{O}\sethlcolor{lightblue}\hl{C}\sethlcolor{darkgreen}\hl{C}\sethlcolor{darkblue}\hl{C}\sethlcolor{darkred}\hl{N}\sethlcolor{lightgreen}\hl{CA}\sethlcolor{lightpink}\hl{C}\sethlcolor{lightyellow}\hl{O}\sethlcolor{lightblue}\hl{C}\sethlcolor{darkgreen}\hl{C}\sethlcolor{darkblue}\hl{N}\sethlcolor{darkred}\hl{C}\sethlcolor{lightgreen}\hl{C}\sethlcolor{lightpink}\hl{N}\sethlcolor{lightyellow}\hl{N}\sethlcolor{lightblue}\hl{CA}\sethlcolor{darkgreen}\hl{C}\sethlcolor{darkblue}\hl{O}\sethlcolor{darkred}\hl{C}\sethlcolor{lightgreen}\hl{C}\sethlcolor{lightpink}\hl{C}\sethlcolor{lightyellow}\hl{C}\sethlcolor{lightblue}\hl{N}\sethlcolor{darkgreen}\hl{CA}\sethlcolor{darkblue}\hl{C}\sethlcolor{darkred}\hl{O}\sethlcolor{lightgreen}\hl{C}\sethlcolor{lightpink}\hl{N}\sethlcolor{lightyellow}\hl{CA}\sethlcolor{lightblue}\hl{C}}};
    \node[below left= 0.5em and 0em of root, anchor=west] (root2) {\texttt{\sethlcolor{lightgreen}\hl{O}\sethlcolor{lightpink}\hl{C}\sethlcolor{lightyellow}\hl{C}\sethlcolor{lightblue}\hl{C}\sethlcolor{darkgreen}\hl{N}\sethlcolor{darkblue}\hl{C}\sethlcolor{darkred}\hl{N}\sethlcolor{lightgreen}\hl{N}\sethlcolor{lightpink}\hl{N}\sethlcolor{lightyellow}\hl{CA}\sethlcolor{lightblue}\hl{C}\sethlcolor{darkgreen}\hl{O}\sethlcolor{darkblue}\hl{C}\sethlcolor{darkred}\hl{C}\sethlcolor{lightgreen}\hl{C}\sethlcolor{lightpink}\hl{N}\sethlcolor{lightyellow}\hl{CA}\sethlcolor{lightblue}\hl{C}\sethlcolor{darkgreen}\hl{O}\sethlcolor{darkblue}\hl{C}\sethlcolor{darkred}\hl{C}\sethlcolor{lightgreen}\hl{O}\sethlcolor{lightpink}\hl{N}\sethlcolor{lightyellow}\hl{N}\sethlcolor{lightblue}\hl{CA}\sethlcolor{darkgreen}\hl{C}\sethlcolor{darkblue}\hl{O}\sethlcolor{darkred}\hl{C}\sethlcolor{lightgreen}\hl{C}\sethlcolor{lightpink}\hl{O}\sethlcolor{lightyellow}\hl{N}\sethlcolor{lightblue}\hl{N}\sethlcolor{darkgreen}\hl{CA}\sethlcolor{darkblue}\hl{C}\sethlcolor{darkred}\hl{O}\sethlcolor{lightgreen}\hl{C}\sethlcolor{lightpink}\hl{C}\sethlcolor{lightyellow}\hl{O}\sethlcolor{lightblue}\hl{O}\sethlcolor{darkgreen}\hl{N}\sethlcolor{darkblue}\hl{CA}\sethlcolor{darkred}\hl{C}\sethlcolor{lightgreen}\hl{O}\sethlcolor{lightpink}\hl{C}\sethlcolor{lightyellow}\hl{C}\sethlcolor{lightblue}\hl{C}\sethlcolor{darkgreen}\hl{C}\sethlcolor{darkblue}\hl{C}\sethlcolor{darkred}\hl{C}\sethlcolor{lightgreen}\hl{C}\sethlcolor{lightpink}\hl{N}\sethlcolor{lightyellow}\hl{CA}\sethlcolor{lightblue}\hl{C}\sethlcolor{darkgreen}\hl{O}\sethlcolor{darkblue}\hl{C}\sethlcolor{darkred}\hl{C}\sethlcolor{lightgreen}\hl{S}\sethlcolor{lightpink}\hl{C}}};
    \node[below left= 0.5em and 0em of root2, anchor=west] (xyz) {\texttt{\sethlcolor{lightyellow}\hl{<XYZ>}}}; 
    \node[below left = 0.5em and 0em of xyz, anchor=west] (c1) {\texttt{\sethlcolor{lightblue}\hl{-4}\sethlcolor{darkgreen}\hl{.991} \phantom{-}\sethlcolor{darkblue}\hl{4}\sethlcolor{darkred}\hl{.794} \phantom{-}\sethlcolor{lightgreen}\hl{6}\sethlcolor{lightpink}\hl{.134}}};
    \node[below left = 0.5em and 0em of c1, anchor=west] (c2) {\texttt{\phantom{-}\sethlcolor{lightyellow}\hl{3}\sethlcolor{lightblue}\hl{.067} \phantom{-}\sethlcolor{darkgreen}\hl{2}\sethlcolor{darkblue}\hl{.185} \sethlcolor{darkred}\hl{-5}\sethlcolor{lightgreen}\hl{.773}}};
    \node[below left = 0.5em and 0em of c2, anchor=west] (c3) {\texttt{\phantom{-}\sethlcolor{lightpink}\hl{0}\sethlcolor{lightyellow}\hl{.121} \sethlcolor{lightblue}\hl{-1}\sethlcolor{darkgreen}\hl{.334} \phantom{-}\sethlcolor{darkblue}\hl{3}\sethlcolor{darkred}\hl{.936}}};
    \node[below left = 0.5em and 0em of c3, anchor=west] (c4) {\texttt{\phantom{-}\sethlcolor{lightgreen}\hl{7}\sethlcolor{lightpink}\hl{.077} \phantom{-}\sethlcolor{lightyellow}\hl{1}\sethlcolor{lightblue}\hl{.335} \phantom{-}\sethlcolor{darkgreen}\hl{2}\sethlcolor{darkblue}\hl{.009}}};
    \node[below left = 0.5em and 0em of c4, anchor=west] (c5) {\texttt{\phantom{-}\sethlcolor{darkred}\hl{1}\sethlcolor{lightgreen}\hl{.134} \sethlcolor{lightpink}\hl{-7}\sethlcolor{lightyellow}\hl{.460} \sethlcolor{lightblue}\hl{-5}\sethlcolor{darkgreen}\hl{.195}}};
    \node[below left = 0.5em and 0em of c5, anchor=west] (c6) {\texttt{\phantom{-}\sethlcolor{darkblue}\hl{2}\sethlcolor{darkred}\hl{.384} \sethlcolor{lightgreen}\hl{-5}\sethlcolor{lightpink}\hl{.084} \phantom{-}\sethlcolor{lightyellow}\hl{0}\sethlcolor{lightblue}\hl{.318}}};
    \node[below left = 0.5em and 0em of c6, anchor=west] (c7) {\texttt{\sethlcolor{darkgreen}\hl{-5}\sethlcolor{darkblue}\hl{.272} \sethlcolor{darkred}\hl{-7}\sethlcolor{lightgreen}\hl{.393} \sethlcolor{lightpink}\hl{-5}\sethlcolor{lightyellow}\hl{.431}}};
    \node[below left = 0.5em and 0em of c7, anchor=west] (c8) {\texttt{\phantom{-}\sethlcolor{lightblue}\hl{7}\sethlcolor{darkgreen}\hl{.391} \sethlcolor{darkblue}\hl{-1}\sethlcolor{darkred}\hl{.954} \phantom{-}\sethlcolor{lightgreen}\hl{0}\sethlcolor{lightpink}\hl{.092}}};
    \node[below left = 0.5em and 0em of c8, anchor=west] (c9) {\texttt{\sethlcolor{lightyellow}\hl{-0}\sethlcolor{lightblue}\hl{.887} \sethlcolor{darkgreen}\hl{-4}\sethlcolor{darkblue}\hl{.613} \phantom{-}\sethlcolor{darkred}\hl{2}\sethlcolor{lightgreen}\hl{.238}}};
    \node[below left = 0.5em and 0em of c9, anchor=west] (c10) {\texttt{\phantom{-}\sethlcolor{lightpink}\hl{0}\sethlcolor{lightyellow}\hl{.488} \phantom{-}\sethlcolor{lightblue}\hl{5}\sethlcolor{darkgreen}\hl{.700} \sethlcolor{darkblue}\hl{-2}\sethlcolor{darkred}\hl{.473}}};
    \node[below left = 0.5em and 0em of c10, anchor=west] (c11) {\texttt{\sethlcolor{lightgreen}\hl{-3}\sethlcolor{lightpink}\hl{.573} \phantom{-}\sethlcolor{lightyellow}\hl{8}\sethlcolor{lightblue}\hl{.085} \phantom{-}\sethlcolor{darkgreen}\hl{4}\sethlcolor{darkblue}\hl{.613}}};
    \node[below left = 0.5em and 0em of c11, anchor=west] (c12) {\texttt{\phantom{-}\sethlcolor{darkred}\hl{3}\sethlcolor{lightgreen}\hl{.905} \sethlcolor{lightpink}\hl{-1}\sethlcolor{lightyellow}\hl{.313} \sethlcolor{lightblue}\hl{-4}\sethlcolor{darkgreen}\hl{.534}}};
    \node[below left = 0.5em and 0em of c12, anchor=west] (c13) {\texttt{\sethlcolor{darkblue}\hl{-10}\sethlcolor{darkred}\hl{.845} \sethlcolor{lightgreen}\hl{7}\sethlcolor{lightpink}\hl{.057} \phantom{-}\sethlcolor{lightyellow}\hl{4}\sethlcolor{lightblue}\hl{.070}}};
    \end{tikzpicture}
    \caption{The visualization of the tokenization for pockets in our language model. We use the same example as in Figure \ref{fig:example_pockets}.}
    \label{fig:example_pockets_vis}
\end{figure}
\newpage

\section{Technical Description of the Training Pipeline}
\label{app:hps}

\subsection{Pretraining}
\label{app:pt_desc}

\begin{wrapfigure}{r}{.55\textwidth}
\vspace{-0.7cm}
\begin{center}
\includegraphics[width=\linewidth]{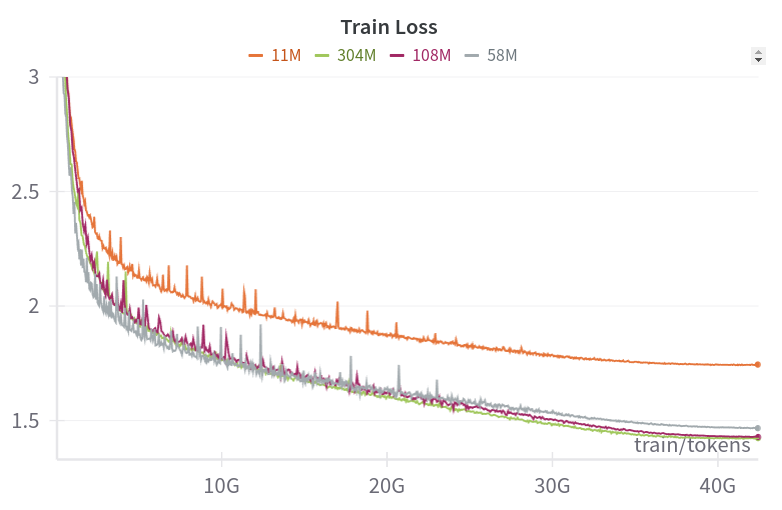}
\end{center}
\vspace{-0.5cm}
\caption{Pretraining loss curves for different model sizes for the representation without explicit hydrogens. }
\vspace{-0.3cm}
\label{fig:pretrain_loss}
\end{wrapfigure}

To achieve efficient pretraining, we use large batch training \citep{keskar2017on} with 1.6M tokens per batch. We set microbatch size to the maximal that fits to the GPU memory and we do gradient accumulation to get large enough batch size (to eventually have 1.6M tokens per batch). Since training sequences have variable length (which comes from the fact that molecules have different sizes), only a part of tokens contribute to the loss so we make sure we have at least 1.6M such ``enabled'' tokens. We use learning rate warmup of 2000 steps, followed by cosine annealing of the learning rate. The maximal learning rate during pre-training is $10^{-3}$ regardless of the model size. We found this many tokens per batch to be important for stable training in this task even with smaller learning rates, especially for models larger than 100M parameters. We use AdamW optimizer \citep{loshchilov2019decoupled} with a weight decay factor of $10^{-2}$. We use gradient clipping with the maximal grad norm of $1.0$. The pretraining takes around 55k optimization steps over 36 hours on one compute node with 8 A6000 GPUs. We employ Flash-Attention2 \cite{dao2023flashattention2} and DeepSpeed optimization accelerator. To use more performant tensor cores, we train with mixed precision where computation is done within the \texttt{bfloat16} datatype. As the distributed optimizer, we use DeepSpeed ZeRO Stage-3 optimizer \citep{rajbhandari2020zero}. We train the model for $1$ epoch only. The amount of tokens in the dataset is $42$B for the version without explicit hydrogens and $90$B tokens for the version with explicit hydrogens. The total size of the Uni-Mol pretraining dataset is around 150GB.

\subsection{Supervised Finetuning}

We use the public CrossDocked version \texttt{v1.3} and for each molecule (except the ones optimized by the Gnina model \citep{McNutt2021} as it yields too many bad itermediate samples) we take its ``minimized'' and ``docked'' formats and extract all intermediate molecules from their files. For each such molecule we cut the pocket with the ProDy \citep{prody} tool. As a result of this process we obtain around 27M pocket-ligand pairs. The size of the CrossDocked that we use is around 50GB.

We use the same recipe for finetuning as for the pretraining with a few changes in hyperparameters. In particular, we use maximal learning rate of $5 \times 10^{-4}$ and only $100$ warump steps. The learning rate schedule, weight decay, optimizer, maximal gradient norm are the same as for the pretraining. The only substantial difference from the pretraining stage is the weighted loss which we use for the CrossDocked finetuning. Specifically, we weight tokens that correspond to different parts of the output, differently. For example, the SMILES tokens have the weight of $1$ while tokens that correspond to the XYZ coordinates placed after SMILES have the weight of $5$. The tokens corresponding to the pocket have the weight of $0$ since they are used as the context only and we don't intend to generate them. As it was described in the Section \ref{sec:sft}, we do SMILES randomization (see \citet{smiles_rand} for implementation details) and rotate pocket it's  ligand randomly - first we sample a random 3D rotation vector, we convert it to a rotation matrix and apply it to the coordinates of both. Also, we enforce the origin of their coordinates to be the same, namely, the coordinate center of the ligand (i.e. we guarantee that the model will generate coordinates around the origin). We train the model on the CrossDocked dataset for $1$ epoch. As it was mentioned in Section \ref{sec:sft}, we extract the full version of the CrossDocked data.

For the finetuning on the GEOM-DRUGS dataset, we use the same hyperparameters as in the SFT stage for CrossDocked with only two differences. First, we weight the loss for all tokens with the same weight of $1$. Second, we don't rotate 3D coordinates of the molecule but only do SMILES randomization. 

\subsection{Reinforcement Learning Funetuning description}
\label{app:rl_desc}

The last stage of our pipeline is Reinforcement Learning. We use a distributed Reinforcement Learning algorithm based on the TRL \citep{vonwerra2022trl} training loop. That is, we launch multiple GPU-workers, where each repeatedly samples experiences, computes rewards, computes the update for the policy (i.e. the transformer language model), synchronizes them, and then performs the gradient update. We use $8$ gpu workers, each with the local batch size of $16$. At every step, we sample a batch of pockets, sample molecules for them, then we compute the rewards via docking tool and perform only one gradient update. We found this to be crucial for our task as otherwise the training might diverge. Even algorithms that are believed to be more powerful, such as PPO \citep{schulman2017proximal}, experience instabilities when the policy lag is bigger. Our surrogate loss for Reinforcement Learning has the following form:

\begin{gather*}
    L(\theta) = \mathbb{E}_{s\sim \mathcal{D}, a\sim p_\theta(a\mid s)} L(\theta, s, a) \\
    L(\theta, s, a) = - R(s, a) \frac{1}{|a|} \log p_{\theta}(a \mid s) + \alpha \text{KL}(p_{\theta_0}(\cdot \mid s) \| p_{\theta}(\cdot\mid s))
\end{gather*}

Here $s$ is the tokenized representation of the pocket and $a$ is the tokenized representation of the generated molecule. $p_{\theta}$ is the current version of the language model being finetuned while $p_{\theta_0}$ is the result of the SFT stage. $\log p_\theta(a \mid s)$ is the sum of generated token log-probabilities. $\mathcal{D}$ is the dataset of prompts (i.e. the pockets-only subset of CrossDocked). $R(s, a)$ is the vina score computed for the corresponding pocket-molecule pair. Finally, we compute the distillation style KL since we want to keep the output distribution of the RL model wide. The KL weight $\alpha$ in our experiments is $\alpha=0.05$. We use a flat learning rate of $1.4 \times 10^{-5}$ and no weight decay. Like before, we clip the gradient norm at $1.0$. In our surrogate loss function, the loglikelihood of the token sequence is averaged (instead of being summed). We found this crucial for training stability.

\section{Evaluating Pretraining at Different Scales}
\label{app:scaling}

\begin{wrapfigure}{r}{.4\textwidth}
\vspace{-0.7cm}
\begin{center}
\includegraphics[width=\linewidth]{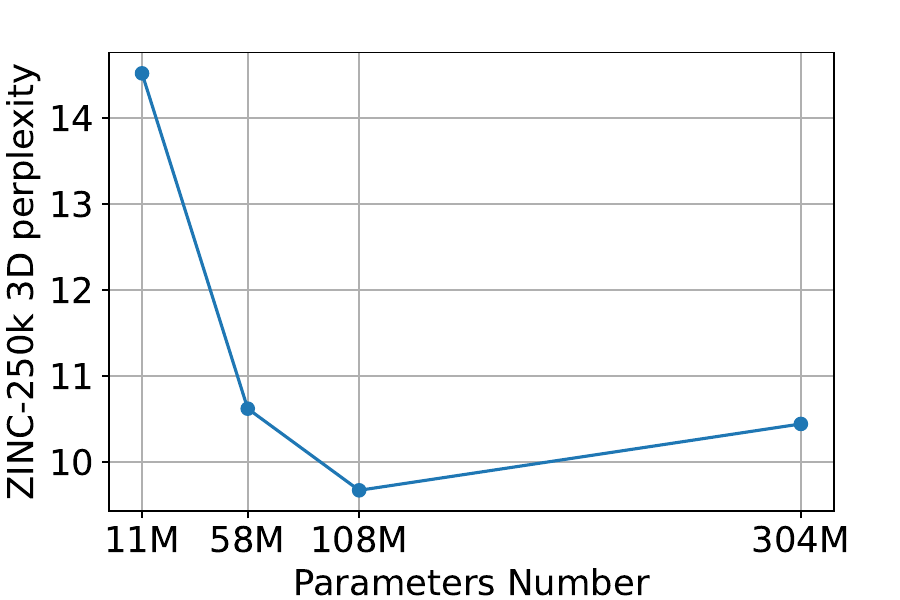}
\end{center}
\vspace{-0.5cm}
\caption{Test perplexity on the hold-out Zinc-250k dataset with 3D conformations from RDKit. BindGPT is the first model capable of this type of evaluation.}
\vspace{-0.3cm}
\label{fig:pretrain_loss}
\end{wrapfigure}

As described in the Section \ref{sec:pretrain}, we pretrain the model on 208M 3D conformations of molecules. By experiment with different models size, we observed that the model scales well up until the size of 300M parameters, where its perplexity shows overfitting. We, therefore, stick to the 100M model in our later experiments as we found it yielding the best results. Figure \ref{fig:pretrain_loss} shows the hold-out test set perplexity on ZINC-250k\footnote{The same procedure was conducted for collecting 3D conformations for ZINC-250k as for the original pretraining data.} for sizes 11M, 58M, 108M, 304M. Note that high value of perplexity is dictated by the highly stochastic nature of 3D molecule coordinates. We believe the model quality can be improved further by increasing the amount of the data for pretraining and the current 108M model obtains the best performance given the pretraining dataset.

\section{RL finetuning Training Curves}
\label{app:rl_finetuning}

\begin{figure}[h!]
    \vspace{-0.4cm}
    \centering
    \subfigure{
        \label{fig:rl_vina}
        \includegraphics[width=0.45\linewidth]{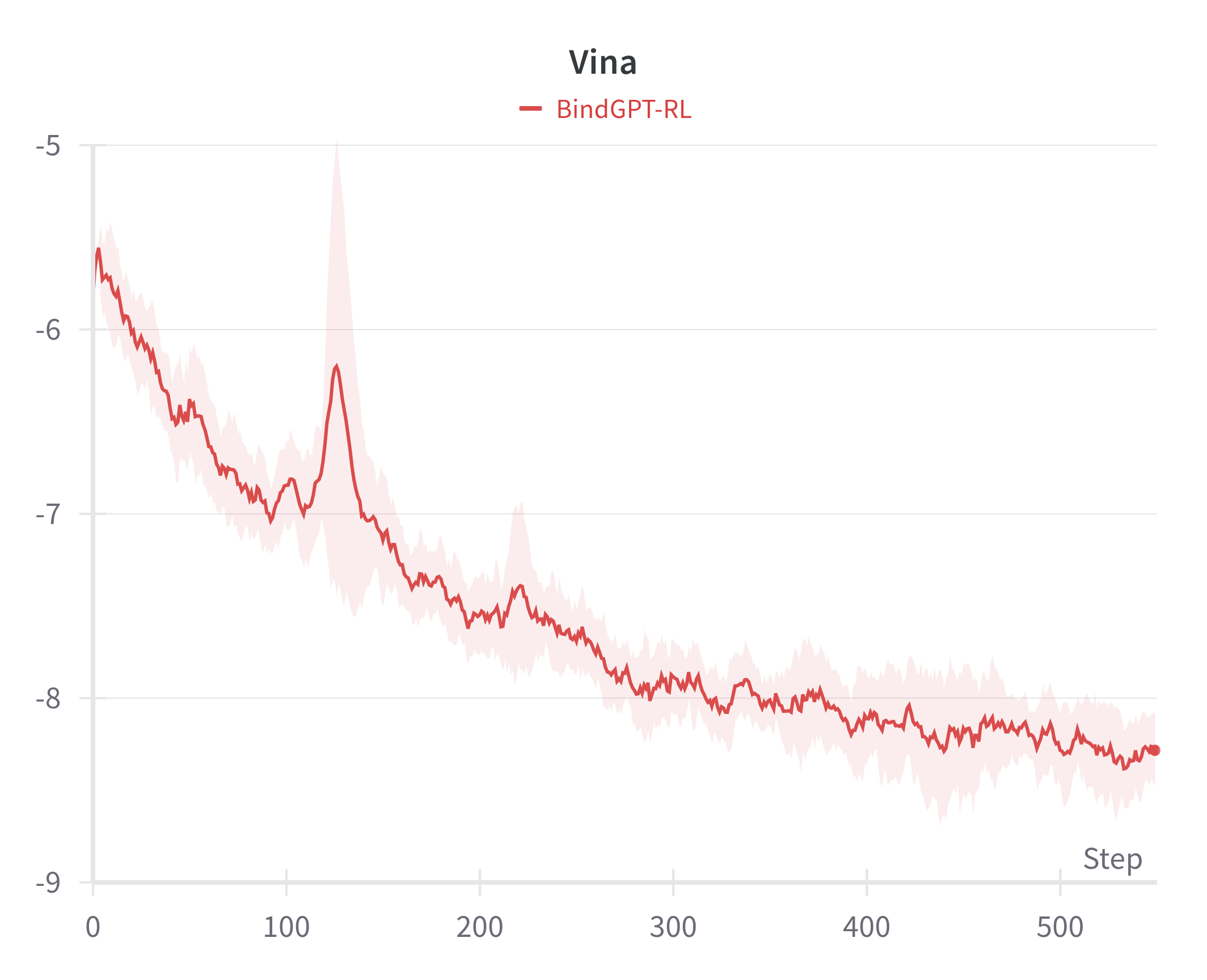}
    }
    \subfigure{
        \label{fig:rl_connected}
        \includegraphics[width=0.45\linewidth]{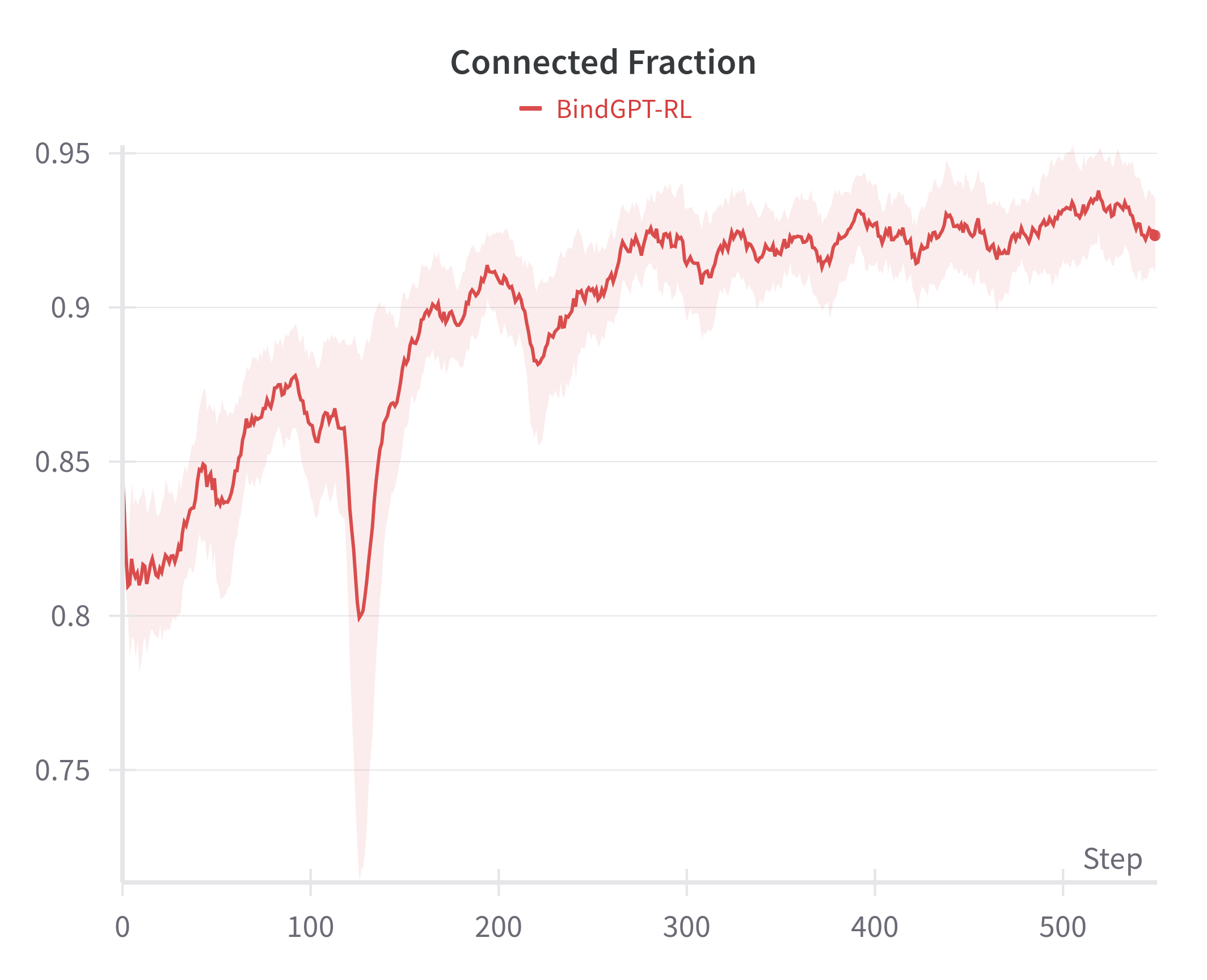}
    }
    
    \subfigure{
        \label{fig:rl_sa}
        \includegraphics[width=0.45\linewidth]{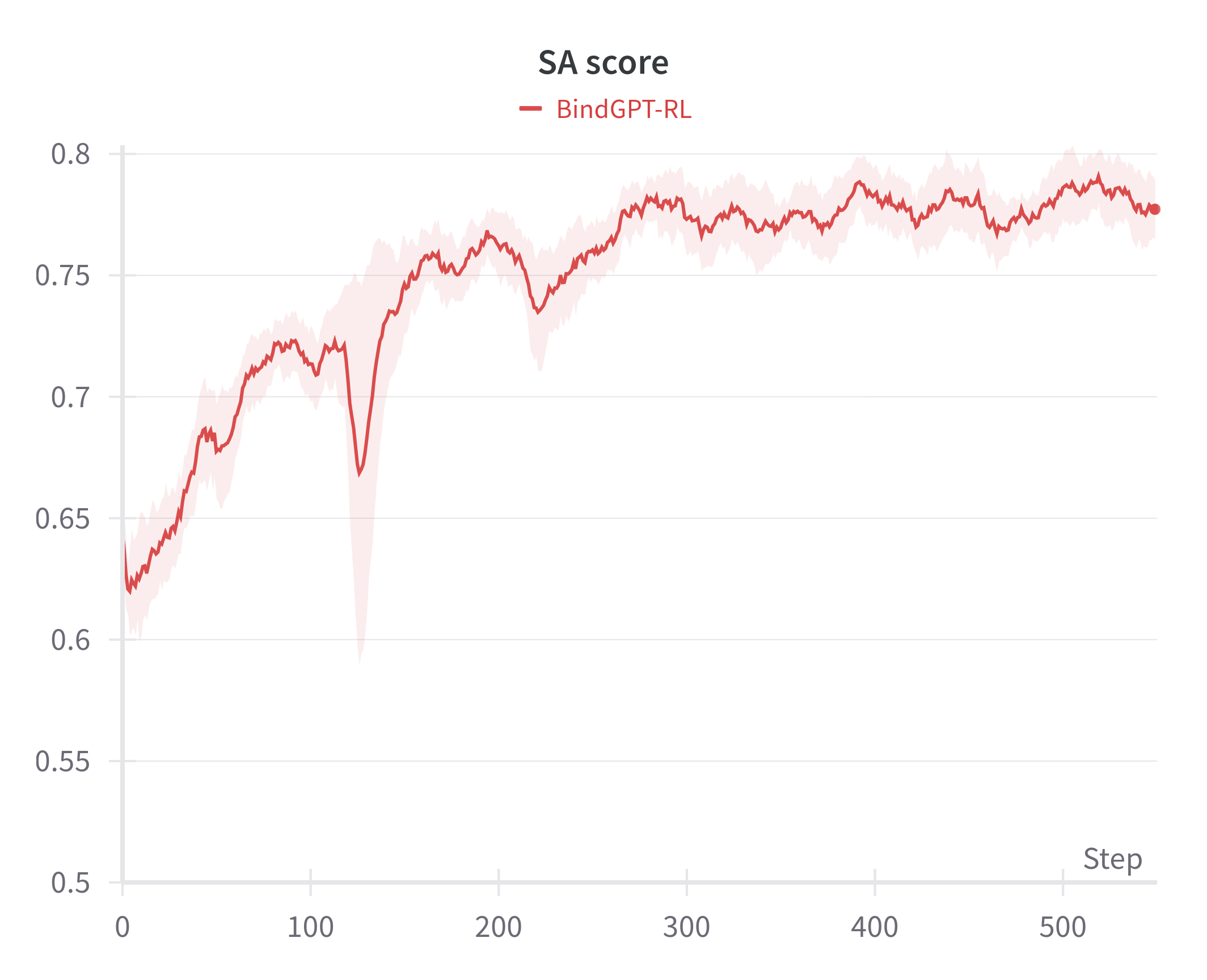}
    }
    \caption{Ligand-pocket affinity objectives averaged over several runs for RL finetuning stage:  {\bf (top left)} Vina Score {\bf (top right)} Ligand connectivity {\bf (bottom)} Synthetic Accessibility}
    \vspace{-0.5cm}
\end{figure}

\newpage

\section{More Samples From the BindGPT Model}
\label{app:more_samples}

\begin{figure}[h!]
    \centering
    \label{fig:pocket_conditioned}
    \includegraphics[width=\linewidth]{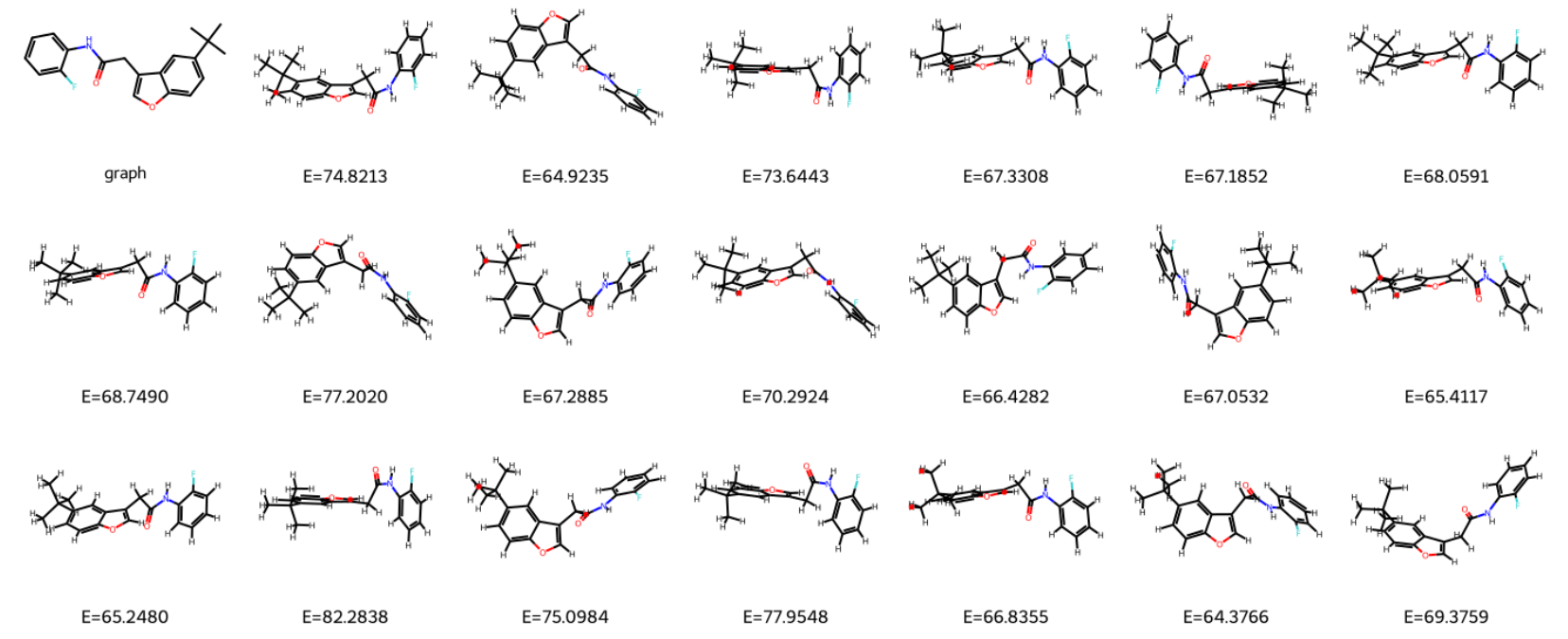}
    \caption{3D conformations generated by BindGPT with explicit hydrogens for a fixed molecule graph. No assistance tools are used. Also, no manual cherry-picking is used here. }
\end{figure}

\begin{figure}[h!]
    \centering
    \label{fig:unconditional_generation}
    \includesvg[width=\linewidth]{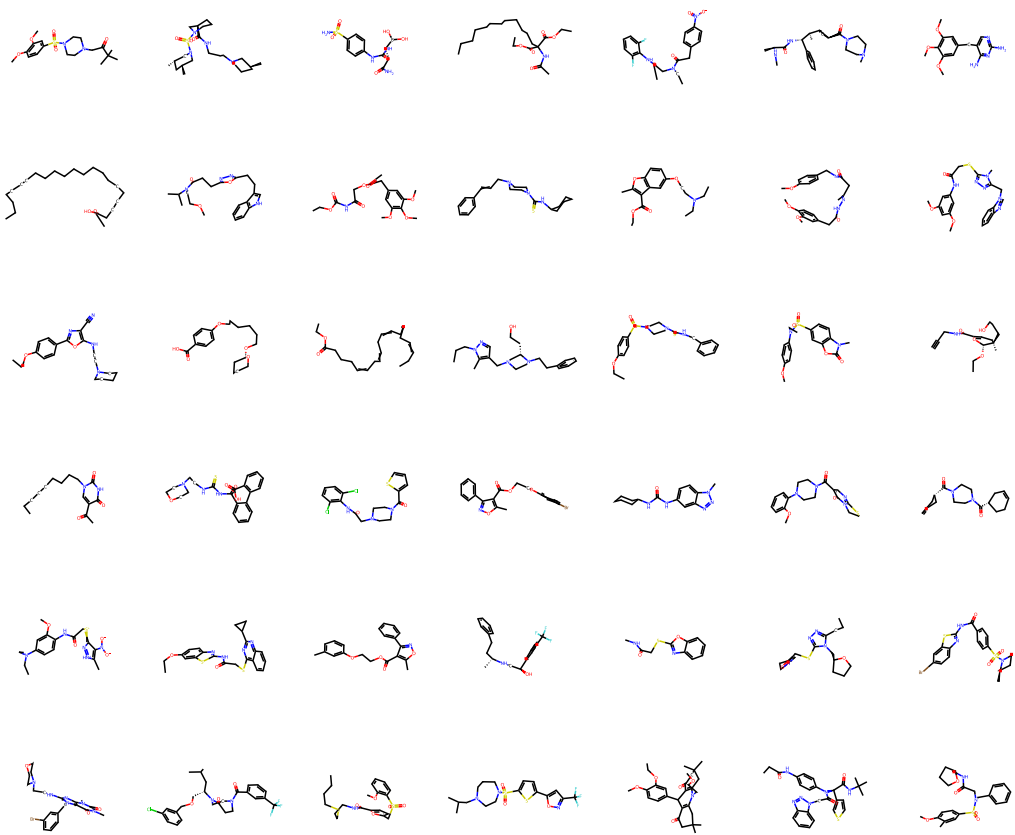}
    \caption{Generated molecules sampled from the model without condition.}
    \vspace{-0.4cm}
\end{figure}

\begin{figure}[h!]
    \centering
    \label{fig:unconditional_generation}
    \includegraphics[width=\linewidth]{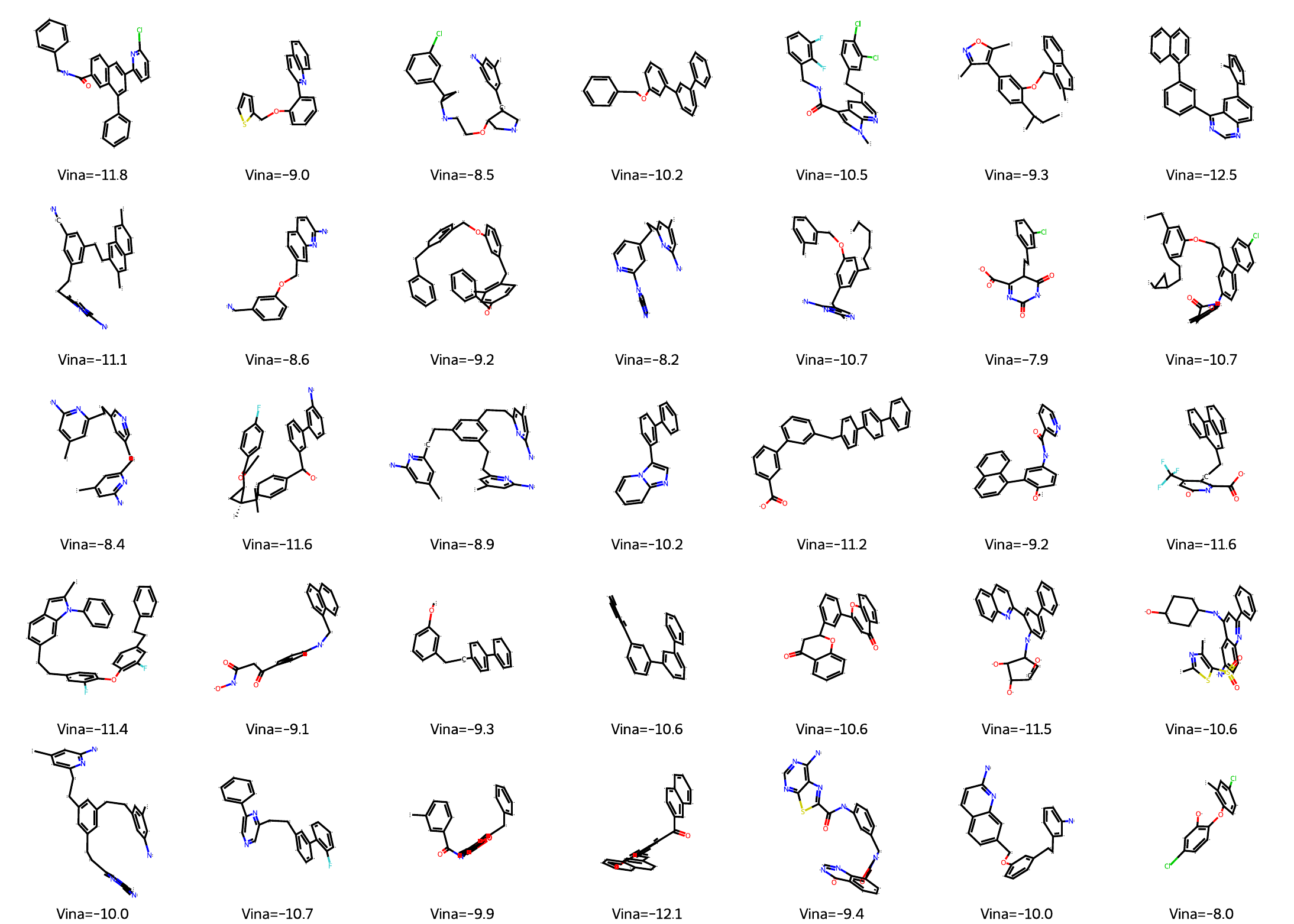}
    \caption{Generated molecules for 4d7o pocket after the RL-finetuning. Note that no cherry-picking (or other filtering) was performed and these are just the first $35$ model samples.}
    \vspace{-0.4cm}
\end{figure}

\section{Augmenting BindGPT with an External Tool for Assisted Generation}
\label{app:augmented_model}

For the unconditional molecule generation tasks and the conformation generation tasks, we enhance the 3D generative abilities of the BindGPT model though the use of the RDKit tool. However, we use it only as a scoring mechanism while our model still acts as the proposal distribution. The scoring happens in the following way: first, we generate SMILES string only (we skip this step if we need to do the conformation generation task). Then, we generate $N$ different conformations from our model and score each with the MMFF \citep{mmff94} energy from RDKit. After that, we select the generated conformation with minimal energy and return it as a sample. It's important to note that we don't use MMFF to optimize the conformation, and also we don't provide the model with any information about the 3D structure. That is, for example, the BindGPT, finetuned on the GEOM-DRUGS will still generate samples within the GEOM conformation distribution (which is different from the distribution produced by MMFF) as can be seen in Table \ref{tab:unconditional_generation}. In addition, when comparing with even higher quality, real world-like conformations, such as the ones of the Platinum dataset, the assisted generation boosts the performance of the model (as can be seen in Figure \ref{fig:conf_gen}), making it's distribution more close to the read world conformations, despite MMFF being just a theoretical approximation. This confirms that such a selection process describes above does not bias the distribution of models outputs, but rather helps it to eliminate generation errors (e.g. atom misplacements). Notably, we don't use the assisted generation for the protein-ligand binding task. In our experiments we use $N=10$.

\section{Efficient Training and Inference}
\label{app:efficient_training}

Despite the wide use of transformers in drug discovery, the majority of current works in this space do not use recent technical advancements that make Large Language Models efficient (for example, Flash Attention \citep{dao2023flashattention2}). The reason is simply that the pretraining paradigm is still coming to drug discovery, with perhaps only one example of the Uni-Mol \citep{zhou2023unimol} model being a multi-task pre-trained transformer model, which unlike this work uses an Encoder-only model that follows the BERT \citep{devlin2019bert} architecture. Therefore, small models are simply trained directly on downstream datasets for which training time optimizations are not crucial. For the case of pretraining, even for a small model, the pretraining over 90B tokens can take a significant time but it obtains a speedup of almost $3\times$ as a result of just using a combination of Flash Attention \citep{dao2022flashattention,dao2023flashattention2} and DeepSpeed \citep{deepspeed}.

To facilitate efficient training and inference, we use the transformers \citep{wolf2020transformers} library from Hugginface with the PyTorch framework \citep{paszke2019pytorch}. We use Flash Attention 2 \citep{dao2023flashattention2} implementation of self-attention, and the DeepSpeed \citep{deepspeed} distributed training framework. During autoregressive sampling, we use Key-Value caching and Flash Attention 2 to speed up decoding. Despite being just implementation optimization tricks, these two techniques can make a big difference for sampling as they speed up decoding by two orders of magnitude compared to the naive approach, making sampling with transformer decoders significantly faster compared to the sampling from diffusion models. 
For example, KV-caching reuses past attention keys and values resulting in $\mathcal{O}(1)$ MLP forward passes instead of $\mathcal{O}(L)$ at each decoding step, where $L$ is the prefix length. Thus, the total number of forward passes with decoding length $L$ is $\mathcal{O}(L)$ instead of $\mathcal{O}(L^2)$.  We hope that our work will promote a more wide use of the transformers best practices within the drug discovery ML community.

\end{document}